\documentclass{article} 
\usepackage{iclr2027_conference,times}

\usepackage{amsmath,amsfonts,bm}

\def\eqref#1{equation~\ref{#1}}

\def\1{\bm{1}}

\DeclareMathAlphabet{\mathsfit}{\encodingdefault}{\sfdefault}{m}{sl}
\SetMathAlphabet{\mathsfit}{bold}{\encodingdefault}{\sfdefault}{bx}{n}

\PassOptionsToPackage{numbers, compress}{natbib}
\usepackage[utf8]{inputenc} 
\usepackage[T1]{fontenc}    
\usepackage{hyperref}       
\usepackage{url}            
\usepackage{booktabs}       
\usepackage{amsfonts}       
\usepackage{nicefrac}       
\usepackage{microtype}      
\usepackage{xcolor}         
\usepackage[table]{xcolor}
\usepackage{multirow}
\usepackage{threeparttable}
\usepackage{makecell}
\usepackage{tabularx}
\usepackage{array}
\usepackage{subcaption}
\usepackage{graphicx}
\usepackage{amsmath}

\title{UniAfford: Token-Routed Multitask Learning for Generalizable 2D-3D Affordance Perception}

\author{%
\parbox{\textwidth}{\centering
  \textbf{Yuhao Liu}\textsuperscript{1,2},
  \textbf{Yiming Zhong}\textsuperscript{1,\(\ddagger\)},
  \textbf{Hanqing Wang}\textsuperscript{4},
  \\[0.25em]
  \textbf{Shaocheng Yan}\textsuperscript{6},
  \textbf{Yuhang Zhang}\textsuperscript{3},
  \textbf{Wenzhou Lyu}\textsuperscript{3},
  \textbf{Ziyang Ding}\textsuperscript{2},
  \textbf{Wei Zhang}\textsuperscript{2},
  \textbf{Xue Zhao}\textsuperscript{3},
  \\[0.25em]
  \textbf{Jin Pan}\textsuperscript{5},
  \textbf{Yuexin Ma}\textsuperscript{1,\(\dagger\)},
  \textbf{Xinge Zhu}\textsuperscript{5}
  \\[0.8em]
  {\small
  \textsuperscript{1}ShanghaiTech University
  \quad
  \textsuperscript{2}Shandong University
  \quad
  \textsuperscript{3}Yinwang Intelligent Technology Co., Ltd.
  \\[0.25em]
  \textsuperscript{4}HKUST(GZ)
  \quad
  \textsuperscript{5}CUHK
  \quad
  \textsuperscript{6}Wuhan University
  }
  \\[0.4em]
  \textsuperscript{\(\dagger\)}Corresponding author
  \quad
  \textsuperscript{\(\ddagger\)}Project Lead
}}

\newif\ifpreprint
\preprinttrue
\makeatletter
\ifpreprint
  \def\@maketitle{%
    \vbox{\hsize\textwidth
      {\LARGE\sc \@title\par}
      \vskip 0.3in
      \begin{center}
        \@author
      \end{center}
      \vskip 0.2in
    }%
  }
\fi
\makeatother

\begin{document}

\maketitle
\lhead{UniAfford}
\rhead{Preprint}
\chead{}

\begin{abstract}
Affordance perception aims to localize actionable regions that support embodied interaction, yet 2D and 3D affordance grounding have evolved as separate problems, with different task definitions, supervision formats, datasets, and evaluation protocols. This fragmentation limits the learning of transferable object--affordance semantics across visual and geometric spaces. We propose \textbf{Token Router for Tasks}, a general multitask training paradigm for MLLM-based systems that routes contextual hidden states to task-specific branches without requiring the language head to generate predefined task markers. Routed states are supervised directly by branch-specific objectives, enabling dense prediction losses to shape shared MLLM representations. We instantiate this paradigm as \textbf{UniAfford}, a unified framework for generalizable 2D--3D affordance perception, together with \textbf{UniAfford-Data}, a unified dataset that integrates pixel-level 2D annotations, point-level 3D annotations, and language instructions under a shared object--affordance taxonomy, supporting heterogeneous supervision through semantic-level 2D--3D pairing. UniAfford adopts an MLLM as a shared semantic hub and a modality-aware token router to produce image- and point-cloud-affordance queries. These queries respectively condition a SAM-style pixel decoder and a SONATA-based point decoder, enabling flexible 2D, 3D, and joint affordance inference from image-only, point-cloud-only, or paired multimodal inputs. Extensive experiments demonstrate strong zero-shot generalization across 2D and 3D affordance benchmarks without target-specific fine-tuning, alongside state-of-the-art branch-wise performance under modality-isolated training and evaluation protocols. Ablations demonstrate the importance of token routing, joint 2D--3D supervision, and decoder coupling, while language-head diagnostics show that routed latent states carry meaningful object--affordance semantics. 
Project page: \href{https://4dvlab.github.io/UniAfford/}{https://4dvlab.github.io/UniAfford/}.
\end{abstract}
\begin{figure*}[t]
    \centering
    \includegraphics[width=0.95\linewidth]{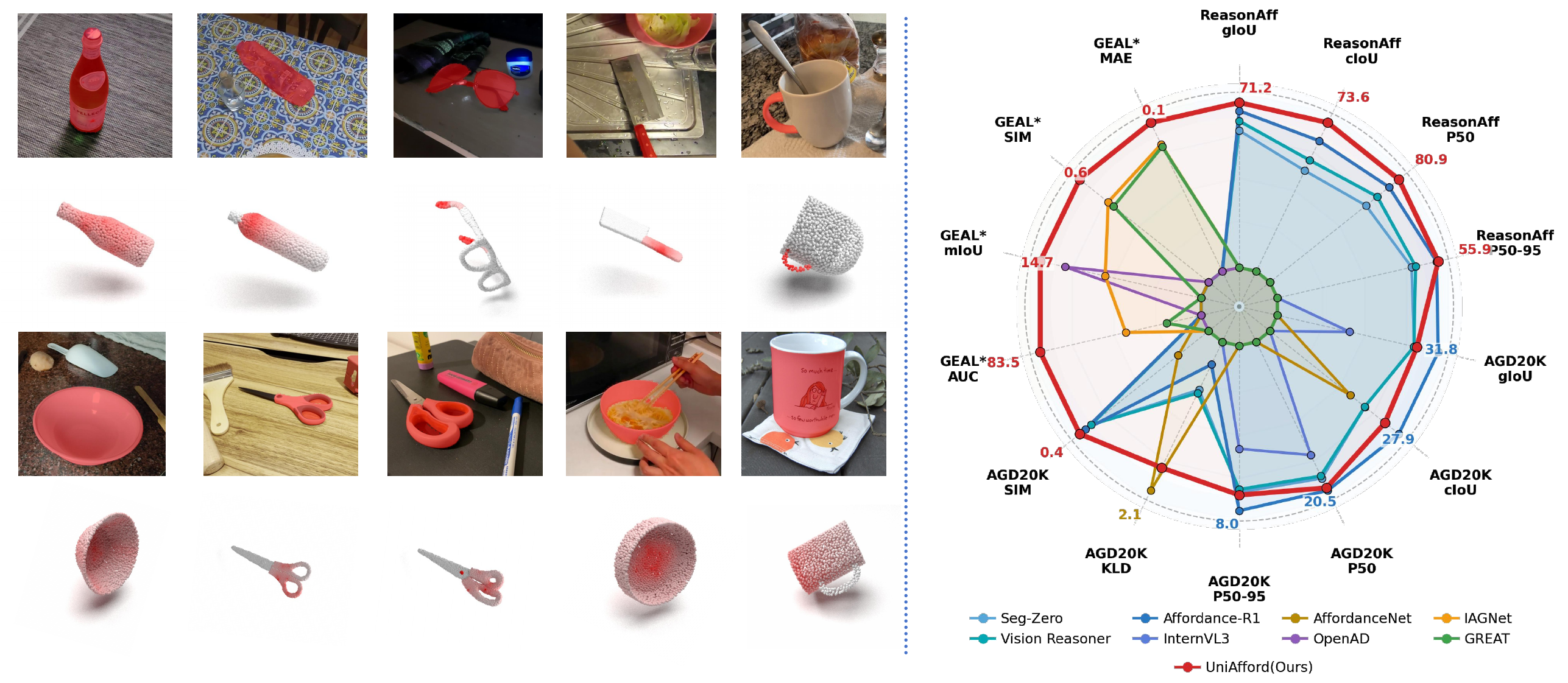}
    \caption{
    \textbf{Overview of UniAfford-Data and UniAfford.}
    Left: Pixel-level and point-level annotations organized under a shared object--affordance taxonomy with semantic pairing across instances.
    Right: UniAfford achieves strong performance across OOD zero-shot transfer and modality-isolated branch-wise evaluations in both 2D and 3D affordance grounding.
    }
    \label{fig:teaser}
    \vspace{-5pt}
\end{figure*}
\section{Introduction}
\label{sec:intro}

As a fundamental capability for embodied intelligence, affordance perception localizes object regions that support interactions, such as grasping a mug, pressing a button, or sitting on a chair. RGB images offer rich appearance and semantic cues, while point clouds provide explicit geometry and physical grounding. We argue that 2D and 3D affordance grounding should be studied as complementary manifestations of a shared affordance understanding problem, rather than isolated modality-specific tasks. A generalizable system should therefore learn transferable object--affordance semantics across visual and geometric spaces and produce actionable predictions in image space, 3D space, or both.

However, existing studies still largely develop these tasks independently. 2D methods benefit from large-scale image data and image-language supervision but lack explicit 3D geometry, whereas 3D methods offer spatial grounding but rely on costly, sparse point-wise annotations. Recent multimodal approaches~\citep{GREAT,IAGNet} combine visual and geometric inputs, yet many still focus on a single output space or use task-specific fusion modules. Such input fusion does not, by itself, unify pixel-level and point-level learning. Consequently, datasets, supervision formats, architectures, and evaluation protocols remain fragmented, limiting cross-modal transfer and zero-shot generalization.

Our goal is not merely to combine two modality-specific predictors, but to jointly learn transferable object--affordance semantics from heterogeneous 2D--3D supervision. A key challenge is that image-only and point-cloud-only samples from different sources may share functional semantics without corresponding to the same physical instance. A unified framework must connect these signals while retaining modality-specific spatial supervision for accurate localization. This raises a central question: \textbf{\emph{can we unify 2D and 3D affordance grounding within a shared framework that learns from heterogeneous supervision and generalizes across benchmarks without target-specific fine-tuning?}} This requires a common semantic organization of data and an interface through which pixel-level and point-level objectives shape shared representations.

MLLMs provide a shared representation space for language, visual, and geometric information, but require an interface for assigning contextual states to downstream branches. Existing methods often dispatch tasks through predefined textual markers and pass corresponding hidden states or text spans to specialized decoders~\citep{UnifiedMLLM,u-llava,next-GPT,LLMBind}. LISA-style methods~\citep{LISA}, for example, use the hidden state of a designated segmentation token whose generation remains language-supervised, coupling task dispatch with predefined token identities. We propose \textbf{Token Router for Tasks}, a general multitask training paradigm that predicts branch assignments directly from contextual MLLM hidden states without requiring the language head to generate predefined task markers. Training-only generic anchors provide route supervision but are excluded from the language modeling objective. Text states are supervised by the language head, while task-routed states are supervised by downstream decoders, enabling heterogeneous dense prediction losses to shape shared representations. At inference, branch assignment is determined directly by the learned router rather than by marker generation.

We instantiate this paradigm as \textbf{UniAfford}, a unified framework for generalizable 2D--3D affordance perception, and construct \textbf{UniAfford-Data}. The dataset integrates pixel-level 2D annotations, point-level 3D annotations, and language instructions under a shared object--affordance taxonomy. It supports image-only, point-cloud-only, and semantically paired multimodal samples, connecting different instances through shared object--affordance labels without requiring instance-level spatial correspondence. UniAfford adopts an MLLM as a shared semantic hub, through which visual and geometric supervision jointly shape affordance representations. A modality-aware router selects and projects contextual states into image- and point-cloud-affordance queries. These queries respectively condition a SAM-style decoder~\citep{SAM} for 2D segmentation and a SONATA-based decoder~\citep{sonata} for 3D point-wise prediction. Shared semantic learning and modality-specific decoding together enable flexible 2D, 3D, and joint affordance inference from available observations.

Trained on UniAfford-Data, UniAfford achieves strong zero-shot generalization on out-of-distribution (OOD) 2D and 3D affordance benchmarks without target-specific fine-tuning. Separate modality-isolated training and evaluation establish state-of-the-art branch-wise performance on the benchmarks. Ablations demonstrate gains from joint 2D--3D supervision, token routing, and decoder coupling, while language-head diagnostics reveal meaningful object--affordance semantics in routed states.
Our contributions are summarized as follows:
\begin{itemize}
    \item We propose \textbf{Token Router for Tasks}, a general multitask training paradigm that decouples task routing from predefined marker generation and enables branch-specific dense prediction objectives to shape shared MLLM representations.

    \item We introduce \textbf{UniAfford} and \textbf{UniAfford-Data} to unify 2D and 3D affordance learning through a shared object--affordance taxonomy, heterogeneous supervision, and semantic-level cross-modal pairing, bridging visual and geometric understanding.

    \item We demonstrate strong OOD zero-shot generalization and state-of-the-art branch-wise performance under modality-isolated training and evaluation protocols. Ablations validate joint 2D--3D supervision, token routing, and similarity-based decoder coupling, while diagnostics reveal object--affordance semantic content in routed states.
\end{itemize}
\section{Related Work}
\label{sec:related}

\subsection{2D and 3D Affordance Grounding}
\label{sec:rw:affordance}

Affordance grounding has developed along two modality-specific directions: image-space grounding and 3D geometric grounding.
2D methods localize functional regions from RGB images through pixel-level segmentation~\citep{AffordanceNet18,multiCNN}, leveraging appearance cues without explicit 3D geometry.
In contrast, 3D methods predict actionable regions on point clouds, providing spatial grounding for robotic interaction~\citep{OpenAD,LASO}, but rely on costly and sparse point-wise annotations.
Although multimodal methods introduce additional visual or linguistic cues, their affordance supervision often remains focused on point-cloud space~\citep{GREAT,IAGNet}.
Despite their shared functional semantics, these directions adopt separate datasets, annotation formats, and evaluation protocols.
UniAfford brings them into a shared affordance learning framework, jointly exploiting pixel-level and point-level supervision under a common object--affordance taxonomy to learn transferable representations across visual and geometric spaces.

\subsection{Language-guided Affordance Grounding}
\label{sec:rw:foundation}

To connect these otherwise separate task spaces, language provides a natural interface for expressing shared object--affordance semantics across modalities.
In 2D vision, LISA~\citep{LISA} links language-conditioned hidden states to mask prediction through its embedding-as-mask interface.
Building on this interface, the AffordanceVLM model accompanying RAGNet~\citep{RAGNet} enables instruction-guided affordance segmentation.
Language guidance similarly connects textual semantics with point-wise affordance prediction in 3D~\citep{OpenAD,LASO,wu2025open}.
DAG~\citep{DAG} leverages affordance priors from text-to-image diffusion models, while SeqAfford~\citep{SeqAfford} uses an MLLM to reason about sequential 3D affordances.
Despite this shared reliance on language, these methods primarily target either image-space masks or point-cloud predictions, leaving pixel-level and point-level supervision largely separate.
UniAfford instead uses language as a common semantic interface for joint 2D--3D affordance learning, with shared MLLM states driving both prediction branches.

\subsection{Cross-modal 2D--3D Learning and Routing}
\label{sec:rw:crossmodal}

While language connects functional semantics across modalities, unified 2D--3D grounding further requires linking shared representations to modality-specific dense predictions.
Cross-modal methods integrate visual and geometric evidence: IAGNet~\citep{IAGNet} transfers interaction cues from images to point-cloud grounding, while GREAT~\citep{GREAT} combines geometric attributes and interaction intentions with visual information.
These approaches emphasize cross-modal fusion for 3D grounding.
Beyond input fusion, joint 2D--3D output supervision raises the question of how shared states should be assigned to different prediction branches, connecting our work to routing architectures and MLLM multitask interfaces.
Mixture-of-experts models~\citep{MoE} use learned gating for sparse expert selection, whereas MLLM multitask frameworks~\citep{UnifiedMLLM,u-llava} connect shared language representations to task-specific components.
Marker-based interfaces, exemplified by LISA~\citep{LISA} and UnifiedMLLM~\citep{UnifiedMLLM}, couple task dispatch with predefined task tokens.
\emph{Token Router for Tasks} instead predicts branch assignments directly from contextual hidden states, decoupling task routing from predefined marker generation.
Text states receive language modeling supervision, while image- and point-cloud-routed states receive branch-specific dense supervision.
This turns shared MLLM representations into a task-structured interface for unified affordance learning from heterogeneous pixel-level and point-level annotations.
\section{Dataset and Task Definition}
\label{sec:dataset}

\subsection{Task Definition}
\label{sec:dataset:task}

We formulate 2D and 3D affordance grounding as a shared semantic task with modality-specific spatial outputs.
Given a language instruction $X$ specifying an object category $o$ and a target affordance category $a$, together with an RGB image $I$, a point cloud $P$, or both, the model predicts
\begin{equation}
    \bigl(\widehat{Y}^{\mathrm{2D}}, \widehat{Y}^{\mathrm{3D}}\bigr)
    = f_{\theta}(X, I, P),
    \label{eq:unified_affordance_task}
\end{equation}
where $\widehat{Y}^{\mathrm{2D}}$ and $\widehat{Y}^{\mathrm{3D}}$ localize the target affordance in image space and point-cloud space, respectively.
Unavailable input modalities and their corresponding outputs are omitted.

A training sample is represented as
$\mathcal{S} = (X, I, P, o, a, Y^{\mathrm{2D}}, Y^{\mathrm{3D}})$,
where $Y^{\mathrm{2D}}$ is a pixel-level affordance mask and $Y^{\mathrm{3D}}$ is a point-level affordance annotation.
Inputs and annotations may be partially available, and each branch receives supervision only when its corresponding observation and annotation are present.
This formulation accommodates image-only, point-cloud-only, and multimodal samples within a shared learning framework.

\subsection{UniAfford-Data}
\label{sec:dataset:construction}

To support unified learning from these heterogeneous annotations, we construct \textbf{UniAfford-Data}, a dataset organized by a shared object--affordance taxonomy.
For 2D supervision, we integrate RGB images with pixel-level affordance masks from RAGNet~\citep{RAGNet} and ReasonAff~\citep{AffordanceR1}.
For 3D supervision, we incorporate point clouds with point-wise affordance annotations from PIADv2~\citep{GREAT} and AGPIL~\citep{AGPIL-LMAffordance3D}.

We normalize object and affordance names across sources into a shared semantic index.
We construct semantic-level pseudo-pairs by matching image and point-cloud instances with the same object--affordance labels.
These pairs connect different instances through shared functional semantics rather than instance-level spatial correspondence.
Each instance retains its original spatial annotation, so the 2D and 3D branches learn to localize the same target affordance on their observations.
This construction makes separate annotation sources jointly usable while preserving modality-specific spatial supervision.
Detailed sourcing and preprocessing are provided in Appendix~\ref{app:dataset:preprocess}.

\begin{table*}[ht]
\centering
\caption{Comparison with  Existing affordance datasets. UniAfford-Data provides both 2D and 3D annotations under a unified object--affordance indexing scheme.}
\vspace{-5pt}
\label{tab:dataset-comparison}
\setlength{\tabcolsep}{4pt}
\resizebox{0.95\linewidth}{!}{
\begin{tabular}{lcccccc}
\toprule
Dataset & Modality & \#Objects & \#Affordances & \#2D Samples & \#3D Samples & Annotation \\
\midrule
UMD~\citep{UMD} & 2D & 17 & 7 & 10k & $\times$ & 2D \\
HANDAL~\citep{HANDAL} & 2D & 17 & 1 & 308k & $\times$ & 2D \\
AGD20K~\citep{AGD20k} & 2D & 50 & 36 & 26k & $\times$ & 2D \\
RAGNet~\citep{RAGNet} & Text, 2D & 180 & -- & 273k & $\times$ & 2D \\
ReasonAff~\citep{AffordanceR1} & Text, 2D & 48 & 30 & 2.5k & $\times$ & Text, 2D \\
\midrule
3D-AffordanceNet~\citep{3dAffordanceNet} & 3D & 23 & 17 & $\times$ & 23k & 3D \\
Affogato~\citep{Affogato} & Text, 3D & $>$450 & $>$350 & $\times$ & 150k & 3D \\
LASO~\citep{LASO} & Text, 3D & 23 & 17 & $\times$ & 19k & 3D \\
GEAL~\citep{GEAL} & Text, 3D & 23 & 17 & $\times$ & 4.8k & 3D \\
AGPIL~\citep{AGPIL-LMAffordance3D} & Text, 2D, 3D & 23 & 17 & 31k & 41k & 3D \\
PIADv2~\citep{GREAT} & Text, 2D, 3D & 43 & 24 & 15k & 38k & 3D \\
\midrule
\textbf{UniAfford-Data (Ours)} & \textbf{Text, 2D, 3D} & \textbf{162} & \textbf{162} & \textbf{25k} & \textbf{69k} & \textbf{2D, 3D} \\
\bottomrule
\end{tabular}
}

\end{table*}

UniAfford-Data retains image-only and point-cloud-only samples alongside semantically paired multimodal samples.
Each record specifies available observations and annotations, determining which branch losses are activated during training.
As summarized in Table~\ref{tab:dataset-comparison}, UniAfford-Data brings pixel-level and point-level annotations into a common object--affordance indexing scheme, providing a unified foundation for learning across visual and geometric spaces.
Additional details on instruction generation, routing-label construction, data splits, and visualizations are provided in Appendix~\ref{app:dataset}.
\section{Method}
\label{sec:method}

\subsection{Overview}
\label{sec:method:overview}

\begin{figure}[h]
    \centering
    \vspace{-5pt}
    \includegraphics[width=\linewidth]{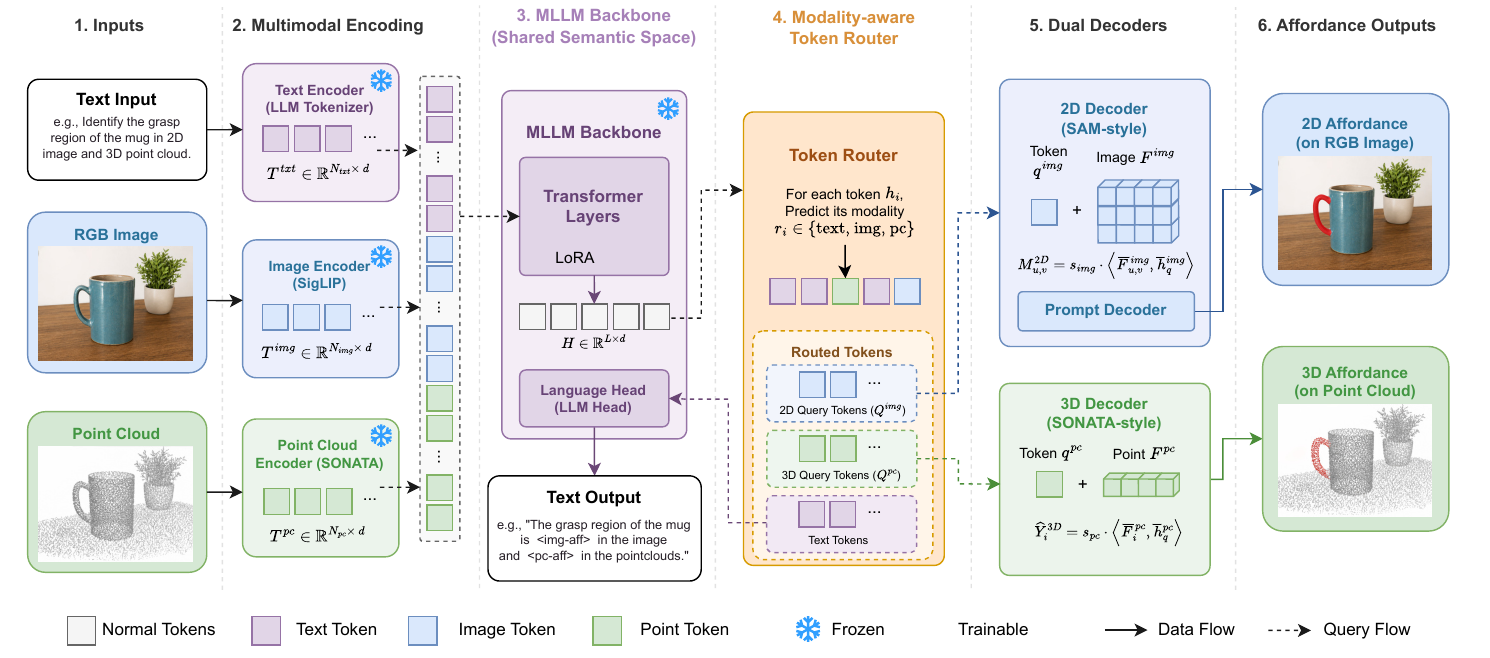}
    \vspace{-10pt}
    \caption{Overview of UniAfford for unified 2D--3D affordance grounding.}
    \vspace{-5pt}
    \label{fig:pipeline}
\end{figure}

We instantiate \emph{Token Router for Tasks} as \textbf{UniAfford}, a unified framework for 2D--3D affordance grounding (Figure~\ref{fig:pipeline}). Given a language instruction and an RGB image, a point cloud, or both, UniAfford uses an MLLM as a shared semantic hub and routes contextual response states to text, 2D affordance, or 3D affordance branches. Routed affordance states condition SAM-style and SONATA-based decoders, respectively, allowing pixel-level and point-level supervision to jointly shape shared representations without requiring the language head to generate predefined task markers.

\subsection{Unified Multimodal Encoding}
\label{sec:method:encoding}

UniAfford maps instruction $X$, image $I$, and point cloud $P \in \mathbb{R}^{N \times 3}$ into a shared MLLM input space through modality-specific encoding pathways for joint semantic processing:
\begin{equation}
    T^{\mathrm{txt}} = E^{\mathrm{mllm}}_{\mathrm{txt}}(X), \qquad
    T^{\mathrm{img}} = E^{\mathrm{mllm}}_{\mathrm{img}}(I), \qquad
    T^{\mathrm{pc}} = E^{\mathrm{mllm}}_{\mathrm{pc}}(P).
\end{equation}
The text pathway performs tokenization, while the image and point-cloud pathways utilize SigLIP and pretrained SONATA encoders followed by projection layers. By dynamically concatenating available modalities, we construct a unified prefix $T^{\mathrm{in}}=[T^{\mathrm{txt}}; T^{\mathrm{img}}; T^{\mathrm{pc}}]$. This prefix conditions the shared MLLM to generate autoregressive responses:
\begin{equation}
    h_t = \mathrm{MLLM}\big(T^{\mathrm{in}},U_{<t}\big)_{\mathrm{last}},
    \qquad H=[h_1;\ldots;h_L],
    \label{eq:mllm_hiddenstate}
\end{equation}
where $U_{<t}$ is the preceding generated sequence and $L$ is the decoding length. The last-layer hidden state $h_t \in \mathbb{R}^{d}$ is utilized for next-token prediction. Subsequently, the Token Router filters valid functional states from $H$, passing them to the downstream decoders for spatial localization.

\subsection{Modality-Aware Token Router}
\label{sec:method:router}

For each valid response state $h_t$, the router predicts a categorical distribution over $\{\mathrm{text},\mathrm{img},\mathrm{pc}\}$:
\begin{equation}
    z_t=g_r(h_t), \qquad
    p_t=\mathrm{softmax}(\widetilde{z}_t),
    \label{eq:router-logits}
\end{equation}
where $g_r$ is a learnable routing head and $\widetilde{z}_t$ masks unavailable branch logits with $-\infty$. A dense branch requires its input and annotation during training, but only its input at inference. Soft probabilities provide differentiable routing supervision, whereas hard assignments select states per branch:
\begin{equation}
    r_t=\arg\max_c p_{t,c}.
    \label{eq:router-labels}
\end{equation}
Branch-specific projections map these states into image- and point-cloud-affordance query spaces:
\begin{equation}
    q^{\mathrm{img}}_t=g_{\mathrm{img}}(h_t), \qquad
    q^{\mathrm{pc}}_t=g_{\mathrm{pc}}(h_t).
\end{equation}
Queries from valid response positions are concatenated in autoregressive order, with padding masks supporting variable-length sequences during batched decoding:
\begin{equation}
    Q^{\mathrm{img}}
    =
    [q_t^{\mathrm{img}} \mid r_t = \mathrm{img}],
    \qquad
    Q^{\mathrm{pc}}
    =
    [q_t^{\mathrm{pc}} \mid r_t = \mathrm{pc}].
\end{equation}
Route labels follow shifted response targets: states whose next targets are \texttt{<img-aff>} or \texttt{<pc-aff>} receive image or point-cloud labels, respectively, while other valid states receive text labels. These anchors are excluded from the language modeling loss and provide only routing supervision; routing-label construction and masking details are provided in Appendix~\ref{app:method:routing}.

\subsection{Affordance Decoders}
\label{sec:method:decoders}

Both decoders couple routed affordance semantics with dense spatial features through similarity-based alignment. Let $q^{\mathrm{img}}$ and $q^{\mathrm{pc}}$ denote individual query representations extracted from the sequences $Q_{\mathrm{img}}$ and $Q_{\mathrm{pc}}$, respectively. For each valid branch $b\in\{\mathrm{img}, \mathrm{pc}\}$, $\phi_b$ and $\gamma_b$ are learnable projections, $s_b$ is a learnable logit scale, and $\mathrm{Norm}$ denotes vector normalization.

\paragraph{2D affordance decoder.}
The SAM~\citep{SAM} encoder extracts dense image features $F^{\mathrm{img}}=E^{\mathrm{dec}}_{\mathrm{img}}(I)$, which are compared with routed image queries through projected similarity to construct a coarse affordance heatmap over the spatial grid:
\begin{equation}
    M^{\mathrm{img}}_{u,v}
    =
    s_{\mathrm{img}}
    \left\langle
    \mathrm{Norm}(\phi_{\mathrm{img}}(F^{\mathrm{img}}_{u,v})),
    \mathrm{Norm}(\gamma_{\mathrm{img}}(q^{\mathrm{img}}))
    \right\rangle,
\end{equation}
where $(u,v)$ indexes the feature grid. The heatmap is resized to the SAM prompt resolution and encoded as a dense mask prompt for spatial refinement:
\begin{equation}
    \widehat{Y}^{\mathrm{2D}}
    =
    D_{\mathrm{2D}}(F^{\mathrm{img}},E^{\mathrm{pmt}}(M^{\mathrm{img}})),
\end{equation}
where $E^{\mathrm{pmt}}$ and $D_{\mathrm{2D}}$ denote the SAM prompt encoder and mask decoder, respectively, and $\widehat{Y}^{\mathrm{2D}}$ contains pixel-wise affordance logits.

\paragraph{3D affordance decoder.}
A SONATA-based~\citep{sonata} encoder extracts dense point features $F^{\mathrm{pc}}=E^{\mathrm{dec}}_{\mathrm{pc}}(P)$ separately from the MLLM-side point encoding. Following the point prompt training philosophy~\citep{ppt}, the decoder computes point-wise affordance logits through scaled query--feature similarity, directly conditioning geometric localization on routed semantics:
\begin{equation}
    \widehat{Y}^{\mathrm{3D}}_i
    =
    s_{\mathrm{pc}}
    \left\langle
    \mathrm{Norm}(\phi_{\mathrm{pc}}(F^{\mathrm{pc}}_i)),
    \mathrm{Norm}(\gamma_{\mathrm{pc}}(q^{\mathrm{pc}}))
    \right\rangle,
    \qquad i=1,\ldots,N.
\end{equation}

\subsection{Training}
\label{sec:method:training}

UniAfford jointly optimizes language modeling, dense prediction, and routing through
$\mathcal{L}=\lambda_{\mathrm{txt}}\mathcal{L}_{\mathrm{txt}}
+m_{\mathrm{2D}}\mathcal{L}_{\mathrm{2D}}+m_{\mathrm{3D}}\mathcal{L}_{\mathrm{3D}}
+\mathcal{L}_\mathrm{router}$,
where $m_{\mathrm{2D}}$ and $m_{\mathrm{3D}}$ indicate input and annotation availability. The language loss supervises ordinary response targets, while dense prediction combines focal and Dice losses for 2D and binary cross-entropy and Dice losses for 3D. The router objective combines token-level cross-entropy with existence and sparsity losses. Dense losses propagate through decoders and query projections into selected MLLM representations, but not through hard routing assignments. The routing head is optimized by $\mathcal{L}_\mathrm{router}$, with routing and dense prediction objectives coupled through shared MLLM representations. Loss definitions and weights are provided in Appendix~\ref{app:method:loss}.
\section{Experiments}
\label{sec:experiments}

We design our experiments to test three core hypotheses behind \textbf{UniAfford}:

\begin{description}
    \item[H1: OOD zero-shot generalization.]
    Can a unified 2D--3D affordance model learn transferable semantics from heterogeneous pixel-level and point-level supervision and generalize to out-of-distribution benchmarks without target-specific training or fine-tuning?

    \item[H2: Branch-wise affordance grounding ability.]
    Can the architecture achieve SOTA or competitive modality-specific performance when branches are trained and evaluated independently?

    \item[H3: Effectiveness of key design choices.]
    Do token routing, joint 2D--3D supervision, and decoder coupling improve affordance grounding, and do routed hidden states carry meaningful object--affordance semantics for pixel-level and point-level prediction?
\end{description}

\subsection{Experimental Setup}
\label{sec:exp:setup}

\paragraph{Evaluation protocols.}
We use two complementary protocols aligned with our hypotheses. The \emph{mixed-training OOD zero-shot} protocol trains UniAfford on UniAfford-Data and directly evaluates it on target benchmarks without target-specific training or fine-tuning, testing cross-dataset transfer of learned object--affordance semantics across visual and geometric spaces. The \emph{modality-isolated} protocol separately trains and evaluates each branch using its target visual modality and task instructions, assessing the standalone grounding capacity of the unified architecture. Dataset choices and protocol-specific settings are detailed in the corresponding subsections.

\paragraph{Metrics and implementation.}
For 2D prediction, we report \textit{gIoU}, \textit{cIoU}, $P_{50}$, and $P_{50\text{--}95}$, with \textit{KLD} and \textit{SIM} additionally used for saliency-style transfer evaluation. For 3D prediction, we report \textit{AUC}, \textit{mIoU}, \textit{SIM}, and \textit{MAE}. We fine-tune the MLLM with LoRA~\citep{LoRA} and optimize using AdamW~\citep{AdamW}. All experiments are conducted on NVIDIA B200 GPUs. Baseline re-evaluations use official implementations and released checkpoints, or models retrained following the corresponding recipes when checkpoints are unavailable. Detailed metric definitions, optimization settings, and baseline implementations are provided in Appendix~\ref{app:exp:setup}.

\subsection{H1: OOD Zero-shot Generalization}
\label{sec:exp:generalization}

We first evaluate whether joint training on UniAfford-Data yields transferable affordance understanding across 2D and 3D benchmarks. For 2D transfer, we evaluate on AGD20K against segmentation-style and reasoning-based MLLM baselines. For 3D transfer, GEAL* denotes a robustness-oriented subset of LASO-C from the GEAL corruption benchmark, comprising Scale, Jitter, and Rotate perturbations at severity level 2. We compare with zero-shot 3D baselines and additionally report LASO~\citep{LASO} and GEAL~\citep{GEAL} trained under the GEAL benchmark setting as references, excluded from the zero-shot ranking.

\begin{table*}[ht]
\centering
\caption{OOD zero-shot transfer results  on unseen 2D and 3D affordance benchmarks.
UniAfford is trained on UniAfford-Data and evaluated on target benchmarks without target-specific training or fine-tuning.
Best zero-shot results are in \textbf{bold}, and second-best zero-shot results are \underline{underlined}.}
\label{tab:mixed-transfer}
\setlength{\tabcolsep}{4pt}

\begin{subtable}[t]{0.58\textwidth}
\centering
\caption{2D transfer results on AGD20K.}
\label{tab:mixed-2d}
\resizebox{\linewidth}{!}{
\begin{tabular}{lccccccc}
\toprule
Model & Reasoning & gIoU$\uparrow$ & cIoU$\uparrow$ & $P_{50\text{--}95}\uparrow$ & $P_{50}\uparrow$ & KLD$\downarrow$ & SIM$\uparrow$ \\
\midrule
Seg-Zero~\citep{Seg-Zero}         & $\checkmark$ & 26.99 & 22.01 & 6.52 & 17.82 & 9.02  & 0.35 \\
Vision Reasoner~\citep{liu2026visionreasonerunifiedreasoningintegratedvisual} 
                                      & $\checkmark$ & 26.98 & 21.98 & 6.31 & 17.31 & 8.90  & 0.35 \\
Affordance-R1~\citep{AffordanceR1} & $\checkmark$ & \textbf{31.78} & \textbf{27.85} & \textbf{7.99} & \textbf{20.49} & 9.73 & \underline{0.36} \\
\midrule
LISA-7B~\citep{LISA}              & $\times$     & 13.18 & 11.96 & 1.45 & 5.31  & 13.68 & 0.16 \\
SAM4MLLM~\citep{SAM4MLLM}         & $\times$     & 15.27 & 13.22 & 2.40 & 6.95  & 9.51  & 0.27 \\
AffordanceNet~\citep{RAGNet} & $\times$ & 14.10 & 20.04 & 0.85 & 3.04  & \textbf{2.10} & 0.26 \\
Qwen2.5VL-7B~\citep{Qwen2.5VL}    & $\times$     & 20.28 & 16.35 & 5.61 & 15.49 & 9.81  & 0.26 \\
InternVL3-7B~\citep{InternVL3}    & $\times$     & 18.18 & 14.63 & 3.79 & 13.37 & 10.09 & 0.25 \\
\midrule
\textbf{UniAfford (Ours)}        & $\times$ & \underline{27.52} & \underline{25.22} & \underline{6.73} & \underline{19.88} & \underline{4.30} & \textbf{0.37} \\
\bottomrule
\end{tabular}
}
\end{subtable}
\hfill
\begin{subtable}[t]{0.38\textwidth}
\centering
\caption{3D transfer results on GEAL*.}
\label{tab:mixed-3d}
\setlength{\tabcolsep}{4pt}
\resizebox{\linewidth}{!}{
\begin{tabular}{lcccc}
\toprule
Method & AUC$\uparrow$ & mIoU$\uparrow$ & SIM$\uparrow$ & MAE$\downarrow$ \\
\midrule
\rowcolor{blue!8}
\multicolumn{5}{l}{\textit{Reference: trained on GEAL}} \\
\midrule
LASO~\citep{LASO} & 81.53 & 16.33 & 0.532 & 0.101 \\
GEAL~\citep{GEAL} & 81.83 & 18.57 & 0.539 & 0.098 \\
\midrule
\rowcolor{purple!8}
\multicolumn{5}{l}{\textit{OOD zero-shot transfer}} \\
\midrule
OpenAD~\citep{OpenAD} & 64.28 & \underline{12.86} & 0.145 & 0.182 \\
IAGNet~\citep{IAGNet} & \underline{69.06} & 10.59 & \underline{0.405} & \underline{0.123} \\
GREAT~\citep{GREAT}  & 65.67 & 8.11  & 0.379 & 0.125 \\
\midrule
\textbf{UniAfford (Ours)} & \textbf{83.55} & \textbf{14.67} & \textbf{0.565} & \textbf{0.102} \\
\bottomrule
\end{tabular}
}
\end{subtable}
\vspace{-4pt}
\end{table*}

Table~\ref{tab:mixed-transfer} demonstrates strong cross-dataset generalization in both output spaces. On AGD20K, UniAfford achieves 27.52 gIoU and 25.22 cIoU, leading all compared methods without explicit reasoning-chain generation on both metrics while remaining competitive with reasoning-based MLLMs. It also achieves the highest SIM of 0.37 among all compared methods. On GEAL*, UniAfford outperforms every evaluated zero-shot baseline across AUC, mIoU, SIM, and MAE, achieving 83.55, 14.67, 0.565, and 0.102, respectively. Despite receiving no target-specific training, it also surpasses the reference models on AUC and SIM. These results establish H1: a unified model trained with heterogeneous pixel-level and point-level supervision achieves strong affordance transfer across modalities and benchmark distributions without target-specific adaptation.

Qualitative comparisons and failure-case analyses on AGD20K and GEAL* are provided in Appendix~\ref{app:qualitative}. These examples illustrate fine-grained functional localization and examine how differences in annotation granularity affect cross-dataset evaluation, complementing the quantitative results with direct comparisons of predicted affordance regions.

\subsection{H2: Branch-wise Affordance Grounding Ability}
\label{sec:exp:benchmark}

We next assess the standalone capability of each branch under separate modality-isolated training and evaluation. The 2D branch uses RGB images and pixel-level masks, while the 3D branch uses point clouds and point-wise annotations. Task instructions remain available in both settings, but UniAfford receives no auxiliary visual observations from the other modality. This protocol examines whether the shared architectural design supports strong grounding in each output space independently.

For 2D grounding, we follow the Affordance-R1~\citep{AffordanceR1} benchmark protocol on ReasonAff. The comparison includes open-vocabulary segmentation models, MLLM-based segmentation models, and reasoning-based affordance models evaluated under the benchmark protocol. For 3D grounding, we train UniAfford on the PIAD or PIADv2 training split and evaluate on PIAD Unseen or PIADv2 Unseen-OBJ, respectively. Object and affordance labels specify the task rather than providing an additional visual modality. Baselines follow their official evaluation settings, retaining auxiliary non-point-cloud cues as required, while UniAfford uses point clouds as its only visual input.

\begin{table*}[h]
\centering
\caption{Branch-wise performance under modality-isolated protocols. 
(a) 2D branch results on ReasonAff. 
(b) 3D branch results under 3D-only training protocols. 
Best results are in \textbf{bold}, and second-best results are \underline{underlined}; for (b), rankings are computed within each training block.}
\label{tab:branch-wise}

\setlength{\tabcolsep}{2.5pt}
\renewcommand{\arraystretch}{0.95}

\begin{subtable}[t]{0.53\textwidth}
\centering
\caption{2D branch on ReasonAff.}
\label{tab:branch-wise-2d}
\resizebox{\linewidth}{!}{
\begin{tabular}{@{}lccccc@{}}
\toprule
Model & Reasoning & gIoU$\uparrow$ & cIoU$\uparrow$ & $P_{50}\uparrow$ & $P_{50\text{--}95}\uparrow$ \\
\midrule
Seg-Zero~\citep{Seg-Zero}              
    & $\checkmark$ & 59.26 & 48.03 & 61.33 & 45.87 \\
Vision Reasoner~\citep{liu2026visionreasonerunifiedreasoningintegratedvisual} 
    & $\checkmark$ & 63.04 & 52.70 & 67.33 & 47.23 \\
Affordance-R1~\citep{AffordanceR1}     
    & $\checkmark$ & \underline{67.41} & \underline{62.72} & \underline{74.50} & \underline{55.22} \\
\midrule
VLPart~\citep{VLPart}                  
    & $\times$ & 4.21  & 3.88  & 1.31  & 0.85 \\
OVSeg~\citep{OVSeg}                    
    & $\times$ & 16.52 & 10.59 & 9.89  & 4.12 \\
SAN~\citep{SAN}                        
    & $\times$ & 10.21 & 13.45 & 7.18  & 3.17 \\
LISA-7B~\citep{LISA}                   
    & $\times$ & 38.17 & 40.58 & 33.62 & 19.69 \\
SAM4MLLM~\citep{SAM4MLLM}              
    & $\times$ & 45.51 & 33.64 & 43.48 & 22.79 \\
AffordanceLLM~\citep{AffordanceLLM}    
    & $\times$ & 48.49 & 38.61 & 42.11 & 20.19 \\
InternVL3-8B~\citep{InternVL3}         
    & $\times$ & 31.79 & 24.68 & 35.41 & 21.93 \\
Qwen2.5VL-7B~\citep{Qwen2.5VL}         
    & $\times$ & 25.18 & 20.54 & 26.00 & 15.82 \\
\midrule
\textbf{UniAfford (Ours)}             
    & $\times$ & \textbf{71.19} & \textbf{73.63} & \textbf{80.94} & \textbf{55.89} \\
\bottomrule
\end{tabular}
}
\end{subtable}
\hfill
\begin{subtable}[t]{0.44\textwidth}
\centering
\caption{3D branch under 3D-only protocols.}
\label{tab:branch-wise-3d}
\resizebox{\linewidth}{!}{
\begin{tabular}{@{}lcccc@{}}
\toprule
Method & AUC$\uparrow$ & mIoU$\uparrow$ & SIM$\uparrow$ & MAE$\downarrow$ \\
\midrule
\rowcolor{blue!8}
\multicolumn{5}{@{}l}{\textit{Trained on PIAD under 3D-only protocol}} \\
\midrule
OpenAD~\citep{OpenAD}                    & 73.75 & 7.810 & 0.384 & 0.125 \\
IAGNet~\citep{IAGNet}                    & 71.84 & 7.950 & 0.352 & 0.127 \\
LASO~\citep{LASO}                        & 71.98 & 8.110 & 0.366 & 0.126 \\
GREAT~\citep{GREAT}                      & 73.61 & 8.820 & 0.384 & 0.124 \\
LMAffordance3D~\citep{AGPIL-LMAffordance3D} 
                                          & 74.02 & 9.050 & \underline{0.390} & 0.127 \\
DAG~\citep{DAG}                          & \underline{76.69} & \underline{9.730} & \textbf{0.414} & \underline{0.120} \\
\midrule
\textbf{UniAfford (Ours)}               & \textbf{77.33} & \textbf{14.25} & \textbf{0.414} & \textbf{0.107} \\
\midrule
\rowcolor{purple!8}
\multicolumn{5}{@{}l}{\textit{Trained on PIADv2 under 3D-only protocol}} \\
\midrule
GREAT~\citep{GREAT}                      & \underline{64.15} & \underline{8.08} & \underline{0.254} & \underline{0.134} \\
\midrule
\textbf{UniAfford (Ours)}               & \textbf{75.67} & \textbf{9.26} & \textbf{0.312} & \textbf{0.127} \\
\bottomrule
\end{tabular}
}
\end{subtable}
 \vspace{-4pt}
\end{table*}

Table~\ref{tab:branch-wise} establishes state-of-the-art branch-wise performance under the evaluated protocols. On ReasonAff, UniAfford leads all reported metrics, improving gIoU from Affordance-R1's 67.41 to 71.19 and cIoU from 62.72 to 73.63. On PIAD, it raises mIoU from DAG's 9.73 to 14.25, an absolute gain of 4.52 points, while achieving the best AUC and MAE and tied-best SIM. On PIADv2, it outperforms GREAT across all four metrics. These results validate H2: the shared MLLM and token-routing architecture delivers strong standalone 2D and 3D grounding, complementing the jointly trained model's cross-dataset generalization in H1.

\subsection{H3: Effectiveness of Key Design Choices}
\label{sec:exp:ablation}

We examine token routing, joint 2D--3D learning, and decoder coupling through controlled ablations. To keep repeated training tractable, all variants use a fixed UniAfford-Data subset and the same held-out partition; single-branch variants use the corresponding modality's training samples. For routing, we replace the learned router with fixed-anchor selection that forwards hidden states associated with \texttt{<img-aff>} and \texttt{<pc-aff>} to their branches, retaining the same backbones and dense decoders. For joint learning, we compare the full model with 2D-only and 3D-only training. For decoder coupling, we replace similarity-based query--feature alignment with prompt-style alternatives.

\begin{table}[h]
\centering
\small
\setlength{\tabcolsep}{4.5pt}
\renewcommand{\arraystretch}{1.12}
\caption{Ablation studies on routing, joint 2D--3D learning, and decoder coupling.
All variants are trained on the same subset of UniAfford-Data and evaluated on the same held-out subset for controlled comparison.
\textbf{Bold} numbers mark the best value for each metric; coupling variants should be interpreted mainly by the branch they modify.}
\label{tab:ablation}
\resizebox{0.9\linewidth}{!}{
\begin{tabular}{ll|cc|cccc}
\toprule
\multirow{2}{*}{Type} & \multirow{2}{*}{Variant}
& \multicolumn{2}{c|}{2D Metrics}
& \multicolumn{4}{c}{3D Metrics} \\
\cmidrule(lr){3-4} \cmidrule(lr){5-8}
& & gIoU$\uparrow$ & cIoU$\uparrow$
& AUC$\uparrow$ & mIoU$\uparrow$ & SIM$\uparrow$ & MAE$\downarrow$ \\
\midrule
Full
& Full model
& \textbf{68.79} & \textbf{58.36}
& \textbf{84.43} & \textbf{34.56} & 0.583 & 0.105 \\
\midrule
Routing
& Fixed-anchor routing
& 62.28 & 55.55
& 84.24 & 17.46 & 0.535 & 0.112 \\
\midrule
\multirow{2}{*}{Joint learning}
& 2D-only training
& 41.41 & 35.74
& -- & -- & -- & -- \\
& 3D-only training
& -- & --
& 82.28 & 30.07 & 0.535 & 0.109 \\
\midrule
\multirow{2}{*}{Coupling}
& Prompt-style 2D coupling
& 36.97 & 23.88
& 74.47 & 22.34 & \textbf{0.589} & \textbf{0.100} \\
& Prompt-style 3D coupling
& 37.48 & 28.26
& 65.33 & 14.31 & 0.417 & 0.160 \\
\bottomrule
\end{tabular}
}
\end{table}

Table~\ref{tab:ablation} demonstrates substantial benefits from learned routing and unified supervision. Compared with fixed-anchor routing, the full model improves 2D gIoU from 62.28 to 68.79 and 3D mIoU from 17.46 to 34.56, establishing the advantage of learned state selection over predefined anchor positions. Joint training raises 2D gIoU from 41.41 to 68.79 and 3D mIoU from 30.07 to 34.56 relative to the respective single-branch variants. Since these variants retain the same branch backbones, this comparison demonstrates the additional value of joint supervision within the shared architecture. These gains directly substantiate the central motivation of UniAfford: heterogeneous pixel-level and point-level supervision jointly improve affordance learning through a shared semantic representation. Replacing similarity-based coupling reduces gIoU to 36.97 for the 2D variant and mIoU to 14.31 for the 3D variant, demonstrating its importance for connecting routed semantics with spatial features. Together, these ablations validate H3 and establish routing, joint supervision, and decoder coupling as key contributors to the framework's performance.

Language-head diagnostics in Appendix~\ref{app:diagnostic} reveal meaningful object and interaction semantics in routed states used for dense prediction. Additional comparisons with the AffordanceNet baseline from RAGNet~\citep{RAGNet} and GREAT~\citep{GREAT}, using their respective modalities from the same ablation subset, further demonstrate UniAfford's performance advantage over specialized models (Appendix~\ref{app:extended_ablation}). Computational profiling in Appendix~\ref{app:cost} reports end-to-end and module-wise costs, including nearly $8\times$ higher 2D throughput than Affordance-R1.

\section{Conclusion}
\label{sec:conclusion}

We presented \emph{Token Router for Tasks}, a general multitask training paradigm that decouples task routing from predefined marker generation, and instantiated it as \textbf{UniAfford} for unified 2D--3D affordance perception. Together with \textbf{UniAfford-Data}, UniAfford jointly learns from image-only, point-cloud-only, and semantically paired samples under a shared object--affordance taxonomy. Routed MLLM states connect shared functional semantics with SAM-style 2D and SONATA-based 3D decoders, allowing pixel-level and point-level supervision to shape common representations. Experiments demonstrate strong cross-dataset zero-shot generalization without target-specific fine-tuning, while modality-isolated training and evaluation establish state-of-the-art branch-wise performance. Ablations validate the benefits of token routing, joint supervision, and similarity-based decoder coupling. UniAfford thus brings 2D and 3D affordance grounding into a shared learning framework for transferable understanding across visual and geometric spaces.

\section{Limitations and Future Work}
\label{sec:limitations}

UniAfford relies on large pretrained backbones, which increases training and inference costs. Semantic-level pairing enables joint learning but does not establish instance-level spatial correspondence, limiting geometric consistency supervision. Our evaluation focuses on affordance grounding benchmarks rather than closed-loop robotic manipulation. Future work will explore more efficient backbone and decoder designs, larger-scale data combining semantic and instance-level pairing, real-world robotic evaluation, and broader multitask dense prediction beyond affordance perception.

\bibliography{iclr2027_conference}

\begin{thebibliography}{42}
\providecommand{\natexlab}[1]{#1}
\providecommand{\url}[1]{\texttt{#1}}
\expandafter\ifx\csname urlstyle\endcsname\relax
  \providecommand{\doi}[1]{doi: #1}\else
  \providecommand{\doi}{doi: \begingroup \urlstyle{rm}\Url}\fi

\bibitem[Bai et~al.(2025)Bai, Chen, Liu, Wang, Ge, Song, Dang, Wang, Wang, Tang, Zhong, Zhu, Yang, Li, Wan, Wang, Ding, Fu, Xu, Ye, Zhang, Xie, Cheng, Zhang, Yang, Xu, and Lin]{Qwen2.5VL}
Shuai Bai, Keqin Chen, Xuejing Liu, Jialin Wang, Wenbin Ge, Sibo Song, Kai Dang, Peng Wang, Shijie Wang, Jun Tang, Humen Zhong, Yuanzhi Zhu, Mingkun Yang, Zhaohai Li, Jianqiang Wan, Pengfei Wang, Wei Ding, Zheren Fu, Yiheng Xu, Jiabo Ye, Xi~Zhang, Tianbao Xie, Zesen Cheng, Hang Zhang, Zhibo Yang, Haiyang Xu, and Junyang Lin.
\newblock Qwen2.5-vl technical report, 2025.
\newblock URL \url{https://arxiv.org/abs/2502.13923}.

\bibitem[Chen et~al.(2024)Chen, Li, Sun, Wang, and Chen]{SAM4MLLM}
Yi-Chia Chen, Wei-Hua Li, Cheng Sun, Yu-Chiang~Frank Wang, and Chu-Song Chen.
\newblock Sam4mllm: Enhance multi-modal large language model for referring expression segmentation, 2024.
\newblock URL \url{https://arxiv.org/abs/2409.10542}.

\bibitem[Do et~al.(2018)Do, Nguyen, and Reid]{AffordanceNet18}
Thanh-Toan Do, Anh Nguyen, and Ian Reid.
\newblock Affordancenet: An end-to-end deep learning approach for object affordance detection.
\newblock In \emph{International Conference on Robotics and Automation (ICRA)}, 2018.

\bibitem[Guo et~al.(2023)Guo, Wen, Yuan, Tremblay, Tyree, Smith, and Birchfield]{HANDAL}
Andrew Guo, Bowen Wen, Jianhe Yuan, Jonathan Tremblay, Stephen Tyree, Jeffrey Smith, and Stan Birchfield.
\newblock {HANDAL}: A dataset of real-world manipulable object categories with pose annotations, affordances, and reconstructions.
\newblock In \emph{IROS}, 2023.

\bibitem[Hu et~al.(2022)Hu, Shen, Wallis, Allen-Zhu, Li, Wang, Wang, and Chen]{LoRA}
Edward~J Hu, Yelong Shen, Phillip Wallis, Zeyuan Allen-Zhu, Yuanzhi Li, Shean Wang, Lu~Wang, and Weizhu Chen.
\newblock Lo{RA}: Low-rank adaptation of large language models.
\newblock In \emph{International Conference on Learning Representations}, 2022.
\newblock URL \url{https://openreview.net/forum?id=nZeVKeeFYf9}.

\bibitem[Jia et~al.(2021)Jia, Xu, Chen, Deng, and Wu]{3dAffordanceNet}
Kui Jia, Xun Xu, Ke~Chen, Shengheng Deng, and Chaozheng Wu.
\newblock 3d affordancenet: A benchmark for visual object affordance understanding, 2021.
\newblock URL \url{https://arxiv.org/abs/2103.16397}.

\bibitem[Kirillov et~al.(2023)Kirillov, Mintun, Ravi, Mao, Rolland, Gustafson, Xiao, Whitehead, Berg, Lo, Dollár, and Girshick]{SAM}
Alexander Kirillov, Eric Mintun, Nikhila Ravi, Hanzi Mao, Chloe Rolland, Laura Gustafson, Tete Xiao, Spencer Whitehead, Alexander~C. Berg, Wan-Yen Lo, Piotr Dollár, and Ross Girshick.
\newblock Segment anything, 2023.
\newblock URL \url{https://arxiv.org/abs/2304.02643}.

\bibitem[Lai et~al.(2024)Lai, Tian, Chen, Li, Yuan, Liu, and Jia]{LISA}
Xin Lai, Zhuotao Tian, Yukang Chen, Yanwei Li, Yuhui Yuan, Shu Liu, and Jiaya Jia.
\newblock Lisa: Reasoning segmentation via large language model.
\newblock In \emph{2024 IEEE/CVF Conference on Computer Vision and Pattern Recognition (CVPR)}, pp.\  9579--9589. IEEE, 2024.

\bibitem[Lee et~al.(2025)Lee, Park, Park, Kang, and Cho]{Affogato}
Junha Lee, Eunha Park, Chunghyun Park, Dahyun Kang, and Minsu Cho.
\newblock Affogato: {Learning} {Open}-{Vocabulary} {Affordance} {Grounding} with {Automated} {Data} {Generation} at {Scale}, 2025.
\newblock URL \url{http://arxiv.org/abs/2506.12009}.

\bibitem[Li et~al.(2024{\natexlab{a}})Li, Zhao, Xiao, Feng, Wang, and Chua]{LASO}
Yicong Li, Na~Zhao, Junbin Xiao, Chun Feng, Xiang Wang, and Tat-seng Chua.
\newblock Laso: Language-guided affordance segmentation on 3d object.
\newblock In \emph{Proceedings of the IEEE/CVF Conference on Computer Vision and Pattern Recognition (CVPR)}, pp.\  14251--14260, June 2024{\natexlab{a}}.

\bibitem[Li et~al.(2024{\natexlab{b}})Li, Wang, Cai, Xu, Wang, Zhang, Song, Jiang, Huang, and Wang]{UnifiedMLLM}
Zhaowei Li, Wei Wang, Yiqing Cai, Qi~Xu, Pengyu Wang, Dong Zhang, Hang Song, Botian Jiang, Zhida Huang, and Tao Wang.
\newblock Unifiedmllm: Enabling unified representation for multi-modal multi-tasks with large language model.
\newblock \emph{ArXiv}, abs/2408.02503, 2024{\natexlab{b}}.

\bibitem[Liang et~al.(2023)Liang, Wu, Dai, Li, Zhao, Zhang, Zhang, Vajda, and Marculescu]{OVSeg}
Feng Liang, Bichen Wu, Xiaoliang Dai, Kunpeng Li, Yinan Zhao, Hang Zhang, Peizhao Zhang, Peter Vajda, and Diana Marculescu.
\newblock Open-vocabulary semantic segmentation with mask-adapted clip, 2023.
\newblock URL \url{https://arxiv.org/abs/2210.04150}.

\bibitem[Liu et~al.(2025{\natexlab{a}})Liu, Wang, Zhang, Chen, Zeng, Ji, Gui, Liu, Yin, and Zhang]{DAG}
Mingyu Liu, Hanqing Wang, Zhenhao Zhang, Yuchao Chen, Xiangyu Zeng, Kaiyang Ji, Tianxiang Gui, Zhirui Liu, Wenti Yin, and Hangxing Zhang.
\newblock Dag: Unleash the potential of diffusion model for open-vocabulary 3d affordance grounding, 2025{\natexlab{a}}.
\newblock URL \url{https://arxiv.org/abs/2508.01651}.

\bibitem[Liu et~al.(2025{\natexlab{b}})Liu, Wang, Zhong, Ma, Wang, Yuan, Cui, Yang, Han, and Wang]{AffordanceR1}
Mingyu Liu, Hanqing Wang, Yiming Zhong, Yuexin Ma, Jiamin Wang, Jiahao Yuan, Zhiqing Cui, Zemin Yang, Yifan Han, and Shaoyang Wang.
\newblock Affordance-r1: Reinforcement learning for generalizable affordance reasoning in multimodal large language model, 2025{\natexlab{b}}.
\newblock URL \url{https://arxiv.org/abs/2508.06206}.

\bibitem[Liu et~al.(2025{\natexlab{c}})Liu, Peng, Zhong, Yue, Lu, Yu, and Jia]{Seg-Zero}
Yuqi Liu, Bohao Peng, Zhisheng Zhong, Zihao Yue, Fanbin Lu, Bei Yu, and Jiaya Jia.
\newblock Seg-zero: Reasoning-chain guided segmentation via cognitive reinforcement, 2025{\natexlab{c}}.
\newblock URL \url{https://arxiv.org/abs/2503.06520}.

\bibitem[Liu et~al.(2026)Liu, Qu, Zhong, Peng, Liu, Yu, and Jia]{liu2026visionreasonerunifiedreasoningintegratedvisual}
Yuqi Liu, Tianyuan Qu, Zhisheng Zhong, Bohao Peng, Shu Liu, Bei Yu, and Jiaya Jia.
\newblock Visionreasoner: Unified reasoning-integrated visual perception via reinforcement learning, 2026.
\newblock URL \url{https://arxiv.org/abs/2505.12081}.

\bibitem[Lobo et~al.(2008)Lobo, Jim{\'e}nez-Valverde, and Real]{AUC}
Jorge~M Lobo, Alberto Jim{\'e}nez-Valverde, and Raimundo Real.
\newblock Auc: a misleading measure of the performance of predictive distribution models.
\newblock \emph{Global ecology and Biogeography}, 17\penalty0 (2):\penalty0 145--151, 2008.

\bibitem[Loshchilov \& Hutter(2017)Loshchilov and Hutter]{AdamW}
Ilya Loshchilov and Frank Hutter.
\newblock Fixing weight decay regularization in adam.
\newblock \emph{CoRR}, abs/1711.05101, 2017.
\newblock URL \url{http://arxiv.org/abs/1711.05101}.

\bibitem[Lu et~al.(2024)Lu, Kong, Huang, and Lee]{GEAL}
Dongyue Lu, Lingdong Kong, Tianxin Huang, and Gim~Hee Lee.
\newblock Geal: Generalizable 3d affordance learning with cross-modal consistency, 2024.
\newblock URL \url{https://arxiv.org/abs/2412.09511}.

\bibitem[Luo et~al.(2022)Luo, Zhai, Zhang, Cao, and Tao]{AGD20k}
Hongchen Luo, Wei Zhai, Jing Zhang, Yang Cao, and Dacheng Tao.
\newblock Learning affordance grounding from exocentric images, 2022.
\newblock URL \url{https://arxiv.org/abs/2203.09905}.

\bibitem[Myers et~al.(2015)Myers, Teo, Ferm\"uller, and Aloimonos]{UMD}
Austin Myers, Ching~L. Teo, Cornelia Ferm\"uller, and Yiannis Aloimonos.
\newblock Affordance detection of tool parts from geometric features.
\newblock In \emph{ICRA}, 2015.

\bibitem[Qian et~al.(2024)Qian, Chen, Bai, Zhou, Tu, and Li]{AffordanceLLM}
Shengyi Qian, Weifeng Chen, Min Bai, Xiong Zhou, Zhuowen Tu, and Li~Erran Li.
\newblock Affordancellm: Grounding affordance from vision language models, 2024.
\newblock URL \url{https://arxiv.org/abs/2401.06341}.

\bibitem[Rahman \& Wang(2016)Rahman and Wang]{mIOU}
Md~Atiqur Rahman and Yang Wang.
\newblock Optimizing intersection-over-union in deep neural networks for image segmentation.
\newblock In \emph{International symposium on visual computing}, pp.\  234--244. Springer, 2016.

\bibitem[Roy \& Todorovic(2016)Roy and Todorovic]{multiCNN}
Anirban Roy and Sinisa Todorovic.
\newblock A multi-scale cnn for affordance segmentation in rgb images.
\newblock In \emph{European conference on computer vision}, pp.\  186--201. Springer, 2016.

\bibitem[Shazeer et~al.(2017)Shazeer, Mirhoseini, Maziarz, Davis, Le, Hinton, and Dean]{MoE}
Noam~M. Shazeer, Azalia Mirhoseini, Krzysztof Maziarz, Andy Davis, Quoc~V. Le, Geoffrey~E. Hinton, and J.~Dean.
\newblock Outrageously large neural networks: The sparsely-gated mixture-of-experts layer.
\newblock \emph{ArXiv}, abs/1701.06538, 2017.

\bibitem[Sun et~al.(2023)Sun, Chen, Zhu, Xiao, Luo, Xie, and Yan]{VLPart}
Peize Sun, Shoufa Chen, Chenchen Zhu, Fanyi Xiao, Ping Luo, Saining Xie, and Zhicheng Yan.
\newblock Going denser with open-vocabulary part segmentation, 2023.
\newblock URL \url{https://arxiv.org/abs/2305.11173}.

\bibitem[Swain \& Ballard(1991)Swain and Ballard]{SIM}
Michael~J Swain and Dana~H Ballard.
\newblock Color indexing.
\newblock \emph{International journal of computer vision}, 7\penalty0 (1):\penalty0 11--32, 1991.

\bibitem[Vo et~al.(2023)Vo, Vu, Huang, Nguyen, Le, Vo, and Nguyen]{OpenAD}
Tuan~Van Vo, Minh~Nhat Vu, Baoru Huang, Toan Nguyen, Ngan Le, Thieu Vo, and Anh Nguyen.
\newblock Open-vocabulary affordance detection using knowledge distillation and text-point correlation, 2023.
\newblock URL \url{https://arxiv.org/abs/2309.10932}.

\bibitem[Willmott \& Matsuura(2005)Willmott and Matsuura]{MAE}
Cort~J Willmott and Kenji Matsuura.
\newblock Advantages of the mean absolute error (mae) over the root mean square error (rmse) in assessing average model performance.
\newblock \emph{Climate research}, 30\penalty0 (1):\penalty0 79--82, 2005.

\bibitem[Wu et~al.(2025{\natexlab{a}})Wu, Fu, Huang, Liu, Jia, Liu, Dai, Wang, Anwer, Khan, and Shen]{RAGNet}
Dongming Wu, Yanping Fu, Saike Huang, Yingfei Liu, Fan Jia, Nian Liu, Feng Dai, Tiancai Wang, Rao~Muhammad Anwer, Fahad~Shahbaz Khan, and Jianbing Shen.
\newblock Ragnet: Large-scale reasoning-based affordance segmentation benchmark towards general grasping, 2025{\natexlab{a}}.
\newblock URL \url{https://arxiv.org/abs/2507.23734}.

\bibitem[Wu et~al.(2025{\natexlab{b}})Wu, Wei, Yu, and Lan]{wu2025open}
Lin Wu, Wei Wei, Peizhuo Yu, and Jianglin Lan.
\newblock Open-vocabulary 3d affordance understanding via functional text enhancement and multilevel representation alignment.
\newblock In \emph{Proceedings of the 33rd ACM International Conference on Multimedia}, pp.\  7988--7997, 2025{\natexlab{b}}.

\bibitem[Wu et~al.(2024{\natexlab{a}})Wu, Fei, Qu, Ji, and Chua]{next-GPT}
Shengqiong Wu, Hao Fei, Leigang Qu, Wei Ji, and Tat-Seng Chua.
\newblock {NE}x{T}-{GPT}: Any-to-any multimodal {LLM}.
\newblock In \emph{Proceedings of the International Conference on Machine Learning}, pp.\  53366--53397, 2024{\natexlab{a}}.

\bibitem[Wu et~al.(2024{\natexlab{b}})Wu, Tian, Wen, Peng, Liu, Yu, and Zhao]{ppt}
Xiaoyang Wu, Zhuotao Tian, Xin Wen, Bohao Peng, Xihui Liu, Kaicheng Yu, and Hengshuang Zhao.
\newblock Towards large-scale 3d representation learning with multi-dataset point prompt training.
\newblock In \emph{CVPR}, 2024{\natexlab{b}}.

\bibitem[Wu et~al.(2025{\natexlab{c}})Wu, DeTone, Frost, Shen, Xie, Yang, Engel, Newcombe, Zhao, and Straub]{sonata}
Xiaoyang Wu, Daniel DeTone, Duncan Frost, Tianwei Shen, Chris Xie, Nan Yang, Jakob Engel, Richard Newcombe, Hengshuang Zhao, and Julian Straub.
\newblock Sonata: Self-supervised learning of reliable point representations.
\newblock In \emph{CVPR}, 2025{\natexlab{c}}.

\bibitem[Xu et~al.(2024)Xu, Xu, Yang, Li, Wang, Xie, Huang, and Li]{u-llava}
Jinjin Xu, Liwu Xu, Yuzhe Yang, Xiang Li, Fanyi Wang, Yanchun Xie, Yi-Jie Huang, and Yaqian Li.
\newblock u-llava: Unifying multi-modal tasks via large language model, 2024.
\newblock URL \url{https://arxiv.org/abs/2311.05348}.

\bibitem[Xu et~al.(2023)Xu, Zhang, Wei, Hu, and Bai]{SAN}
Mengde Xu, Zheng Zhang, Fangyun Wei, Han Hu, and Xiang Bai.
\newblock Side adapter network for open-vocabulary semantic segmentation, 2023.
\newblock URL \url{https://arxiv.org/abs/2302.12242}.

\bibitem[Yang et~al.(2023)Yang, Zhai, Luo, Cao, Luo, and Zha]{IAGNet}
Yuhang Yang, Wei Zhai, Hongchen Luo, Yang Cao, Jiebo Luo, and Zheng-Jun Zha.
\newblock Grounding 3d object affordance from 2d interactions in images.
\newblock In \emph{Proceedings of the IEEE/CVF International Conference on Computer Vision (ICCV)}, pp.\  10905--10915, October 2023.

\bibitem[Yang et~al.(2024)Yang, Zhai, Luo, Cao, Zha, and Shao]{GREAT}
Yuhang Yang, Wei Zhai, Hongchen Luo, Yang Cao, Zheng-Jun Zha, and Yawen Shao.
\newblock Great: Geometry-intention collaborative inference for open-vocabulary 3d object affordance grounding, 2024.
\newblock URL \url{https://arxiv.org/abs/2411.19626}.

\bibitem[Yu et~al.(2025)Yu, Wang, Shi, Luo, Yang, Yu, and Wang]{SeqAfford}
Chunlin Yu, Hanqing Wang, Ye~Shi, Haoyang Luo, Sibei Yang, Jingyi Yu, and Jingya Wang.
\newblock Seqafford: Sequential 3d affordance reasoning via multimodal large language model, 2025.
\newblock URL \url{https://arxiv.org/abs/2412.01550}.

\bibitem[Zhu et~al.(2026)Zhu, Ning, Jin, Lin, Huang, Song, Zhang, Tang, Pan, and Yuan]{LLMBind}
Bin Zhu, Munan Ning, Peng Jin, Bin Lin, Jinfa Huang, Qi~Song, Junwu Zhang, Zhenyu Tang, Mingjun Pan, and Li~Yuan.
\newblock Llmbind: A unified modality-task integration framework, 2026.
\newblock URL \url{https://arxiv.org/abs/2402.14891}.

\bibitem[Zhu et~al.(2025{\natexlab{a}})Zhu, Kong, Xu, Xia, Deng, Ye, Xiong, and Wang]{AGPIL-LMAffordance3D}
He~Zhu, Quyu Kong, Kechun Xu, Xunlong Xia, Bing Deng, Jieping Ye, Rong Xiong, and Yue Wang.
\newblock Grounding 3d object affordance with language instructions, visual observations and interactions, 2025{\natexlab{a}}.
\newblock URL \url{https://arxiv.org/abs/2504.04744}.

\bibitem[Zhu et~al.(2025{\natexlab{b}})Zhu, Wang, Chen, Liu, Ye, Gu, Tian, Duan, Su, Shao, Gao, Cui, Wang, Cao, Liu, Wei, Zhang, Wang, Xu, Li, Wang, Deng, Li, He, Jiang, Luo, Wang, He, Shi, Zhang, Shao, He, Xiong, Qu, Sun, Jiao, Lv, Wu, Zhang, Deng, Ge, Chen, Wang, Dou, Lu, Zhu, Lu, Lin, Qiao, Dai, and Wang]{InternVL3}
Jinguo Zhu, Weiyun Wang, Zhe Chen, Zhaoyang Liu, Shenglong Ye, Lixin Gu, Hao Tian, Yuchen Duan, Weijie Su, Jie Shao, Zhangwei Gao, Erfei Cui, Xuehui Wang, Yue Cao, Yangzhou Liu, Xingguang Wei, Hongjie Zhang, Haomin Wang, Weiye Xu, Hao Li, Jiahao Wang, Nianchen Deng, Songze Li, Yinan He, Tan Jiang, Jiapeng Luo, Yi~Wang, Conghui He, Botian Shi, Xingcheng Zhang, Wenqi Shao, Junjun He, Yingtong Xiong, Wenwen Qu, Peng Sun, Penglong Jiao, Han Lv, Lijun Wu, Kaipeng Zhang, Huipeng Deng, Jiaye Ge, Kai Chen, Limin Wang, Min Dou, Lewei Lu, Xizhou Zhu, Tong Lu, Dahua Lin, Yu~Qiao, Jifeng Dai, and Wenhai Wang.
\newblock Internvl3: Exploring advanced training and test-time recipes for open-source multimodal models, 2025{\natexlab{b}}.
\newblock URL \url{https://arxiv.org/abs/2504.10479}.

\end{thebibliography}
\bibliographystyle{iclr2027_conference}

\newpage

\appendix
\section{Additional Details of UniAfford-Data}
\label{app:dataset}

\subsection{Unified Data Layout and Indexing}
\label{app:dataset:schema}

UniAfford-Data organizes language instructions, RGB images, and point clouds within a unified object-centric directory structure. Each object directory contains an instruction table and optional image and point-cloud annotations. Normalized object and affordance names define a shared semantic index, while modality-specific sample identifiers distinguish individual observations.

The instruction table records optional \texttt{img\_id} and \texttt{pc\_id} bindings, allowing image-only, point-cloud-only, and semantically paired multimodal samples to share the same loader. These bindings associate observations with a common grounding task; they do not imply that the image and point cloud depict the same physical instance. Table~\ref{tab:app-metadata-schema} summarizes the storage layout.

\begin{table}[ht]
    \centering
    \caption{
    Storage layout of UniAfford-Data. Semantic labels provide a shared
    index, while sample identifiers link instructions to the corresponding
    observations and annotations.
    }
    \label{tab:app-metadata-schema}
    \begin{tabular}{@{}p{0.28\linewidth}p{0.66\linewidth}@{}}
        \toprule
        Component & Stored content \\
        \midrule
        \texttt{Instruction.csv}
        & Language instructions, object and affordance labels, and
        optional \texttt{img\_id}/\texttt{pc\_id} bindings \\

        \texttt{Image/}
        & RGB images and pixel-level affordance masks grouped by
        affordance label \\

        \texttt{PointCloud/}
        & CSV point clouds with \texttt{x,y,z} coordinates followed
        by point-wise affordance labels \\

        \texttt{train.json}, \texttt{val.json}, \texttt{test.json}
        & Split-specific instruction and modality-sample identifiers
        indexed by object and affordance \\
        \bottomrule
    \end{tabular}
\end{table}

\subsection{Data Sourcing and Preprocessing}
\label{app:dataset:preprocess}

\paragraph{Data sources.}
UniAfford-Data integrates existing annotations from both image-space and point-cloud-space affordance datasets. For 2D supervision, we use RGB images and pixel-level affordance masks primarily from RAGNet~\citep{RAGNet} and ReasonAff~\citep{AffordanceR1}. For 3D supervision, we incorporate point clouds and point-wise affordance annotations from PIADv2~\citep{GREAT} and AGPIL~\citep{AGPIL-LMAffordance3D}. The source annotations are retained, while their storage formats and semantic labels are organized for joint training.

\paragraph{Spatial preprocessing.}
RGB images and their corresponding masks are resized to a common spatial resolution, with target mask values normalized to $[0,1]$. Point clouds are sampled to a fixed number of points and normalized in scale. The same sampling indices are applied to coordinates and point-wise labels, preserving the correspondence between each sampled point and its affordance annotation. Image resolution and point counts follow the protocol-specific settings described in Appendix~\ref{app:exp:implementation}.

\paragraph{Taxonomy-aligned semantic pseudo-pairing.}
We normalize object categories and affordance labels across sources into a shared semantic index. Semantic pseudo-pairs are then constructed dynamically by associating image and point-cloud instances with the same normalized object--affordance combination. Matching therefore requires agreement on both the object category and the target affordance, rather than the object category alone.

For an object--affordance combination $(o,a)$, let $(I_i,Y_i^{\mathrm{2D}})$ and $(P_j,Y_j^{\mathrm{3D}})$ denote annotated image and point-cloud instances indexed by these labels. A semantically paired training sample is represented as
\begin{equation}
    \mathcal{S}^{\mathrm{pair}}_{ij}
    =
    \bigl(
    X(o,a), I_i, P_j, o, a,
    Y_i^{\mathrm{2D}}, Y_j^{\mathrm{3D}}
    \bigr),
    \label{eq:app:semantic-pair}
\end{equation}
where $X(o,a)$ is a shared instruction instantiated from the object and affordance labels. The two observations represent different physical instances and need not share geometry, viewpoint, or spatial coordinates.

\paragraph{Preserving instance-specific supervision.}
Each observation retains its source spatial annotation: the 2D prediction is supervised by $Y_i^{\mathrm{2D}}$, and the 3D prediction is supervised by $Y_j^{\mathrm{3D}}$. Consequently, semantic pairing does not transfer masks or point labels between instances or impose pixel-to-point correspondence. Differences in shape, viewpoint, and annotation extent remain associated with the respective observations.

This construction separates semantic unification from geometric correspondence. It makes heterogeneous annotation sources jointly usable without requiring newly captured instance-aligned 2D--3D data, while preserving the spatial supervision needed by each prediction branch. Image-only and point-cloud-only records remain available alongside semantic pseudo-pairs, allowing the framework to learn from both individual modalities and their shared functional semantics.

\subsection{Instruction Generation and Routing Supervision}
\label{app:dataset:template}

Each sample requires a textual instruction specifying the target object and affordance. For source examples with human-written descriptions, we preserve the original instruction. When no such description is provided, we dynamically construct a two-part dialogue template based on the normalized object category $obj$ and affordance category $aff$. Crucially, this template strictly separates the condition prompt from the supervised training target:
\begin{quote}
    \textbf{User Query:} ``Locate the \{$aff$\} affordance region of the \{$obj$\}.''\\
    \textbf{Answer:} ``In the 2D image, the \{$aff$\} affordance region of the \{$obj$\} is \texttt{<img-aff>}; in the 3D point cloud, it is \texttt{<pc-aff>}.''
\end{quote}
The \textbf{User Query} serves as the input instruction, while the \textbf{Answer} template provides ordinary language targets together with anchor-derived routing supervision. For semantic pseudo-pairs, this shared instruction template binds the independent image and point-cloud observations to the same task. The textual input specifies the functional objective, while the corresponding pixel- and point-level annotations determine its dense spatial realization on each visual instance.

\paragraph{Routing-Label Construction and Loss Masking.} 
To seamlessly connect the textual response with dense predictions, the assistant response utilizes generic routing anchors (i.e., \texttt{<img-aff>} and \texttt{<pc-aff>}). During training, route labels $y_t^{\mathrm{route}}$ follow shifted response targets: a valid response state receives an image-route or point-cloud-route target label when its next target token is \texttt{<img-aff>} or \texttt{<pc-aff>}, respectively. Other valid response states receive the text-route label. It is important to note that the exact positions of these \texttt{<img-aff>} and \texttt{<pc-aff>} anchors are explicitly excluded and skipped from the ordinary language modeling loss. Since they specify branch roles rather than standard vocabulary targets, their positions provide solely routing supervision. Actual branch availability (and thus dense supervision) is governed dynamically by the loaded paired observations (e.g., whether a 2D image and/or a 3D point cloud is actively available). Further details regarding this routing-label supervision and masking strategy are provided in Appendix~\ref{app:method:routing}.

\subsection{Data Splits}
\label{app:dataset:splits}

Dataset partitions are stored in \texttt{train.json}, \texttt{val.json}, and \texttt{test.json}. Each split uses a modality-specific object--affordance index of the form
\begin{center}
    \texttt{\{Instruction, Image, PointCloud\}}
    $\rightarrow$ object
    $\rightarrow$ affordance
    $\rightarrow$ sample ids.
\end{center}
For \texttt{Instruction}, an entry is either an instruction identifier or a record containing \texttt{id} and optional \texttt{img\_id}/\texttt{pc\_id} bindings. For \texttt{Image} and \texttt{PointCloud}, entries identify modality-specific samples. The semantic index organizes the grounding tasks, while the sample identifiers track the observations and annotations associated with each record.

We provide two dataset versions for experiments at different scales. \texttt{Sample} is a lightweight subset with train, validation, and test partitions for rapid prototyping, branch-wise debugging, and ablation studies. \texttt{Final} is the full-scale version used for large-scale training and validation. These version names distinguish dataset scale; the modality-isolated experiments instead follow the corresponding benchmark protocols described in Section~\ref{sec:exp:benchmark}.

%

\subsection{Dataset Visualization}
\label{app:dataset:visualization}

\begin{figure}[t]
    \centering
    \includegraphics[width=0.92\linewidth]{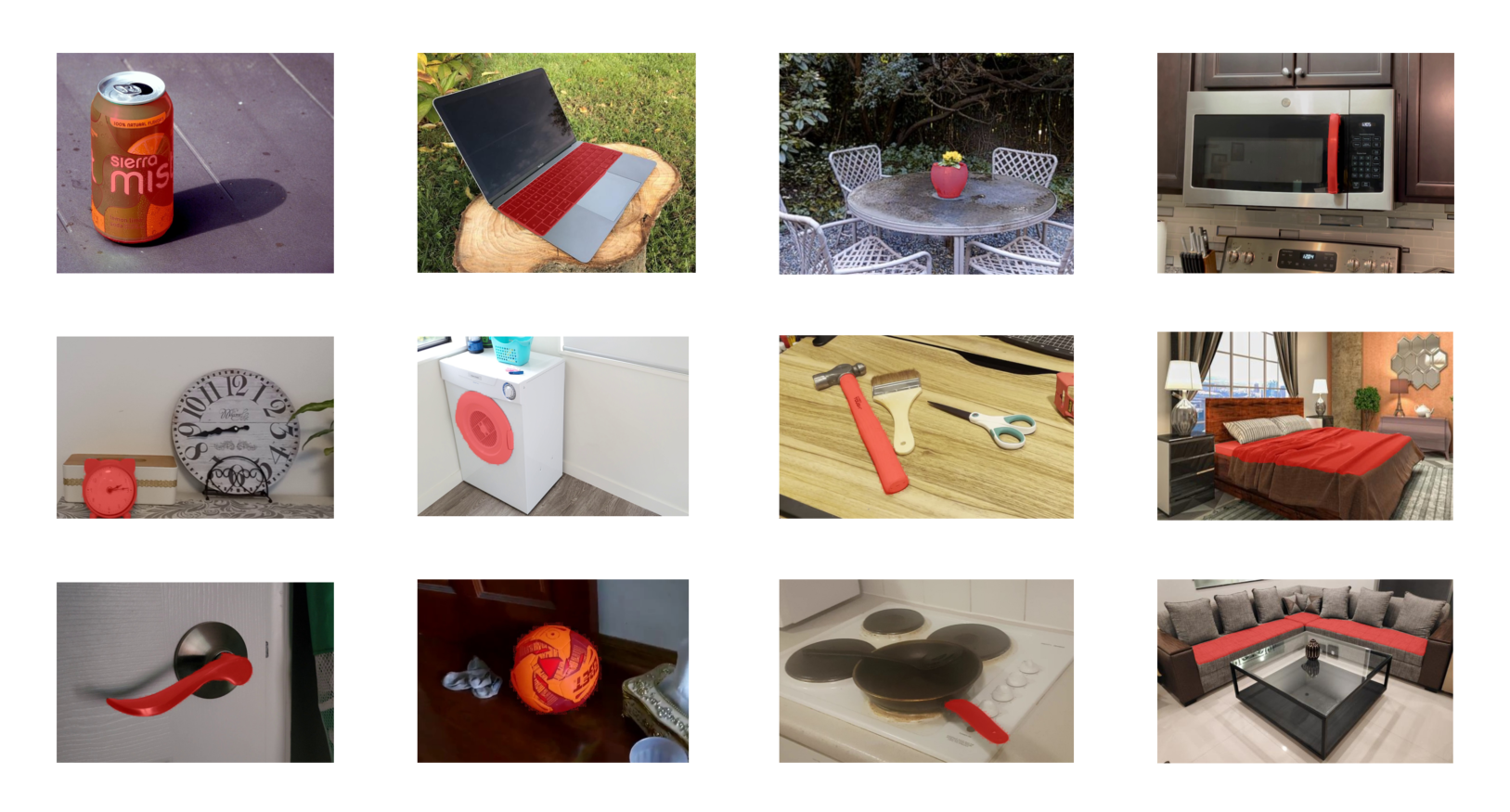}\\[0.8em]
    \includegraphics[width=0.92\linewidth]{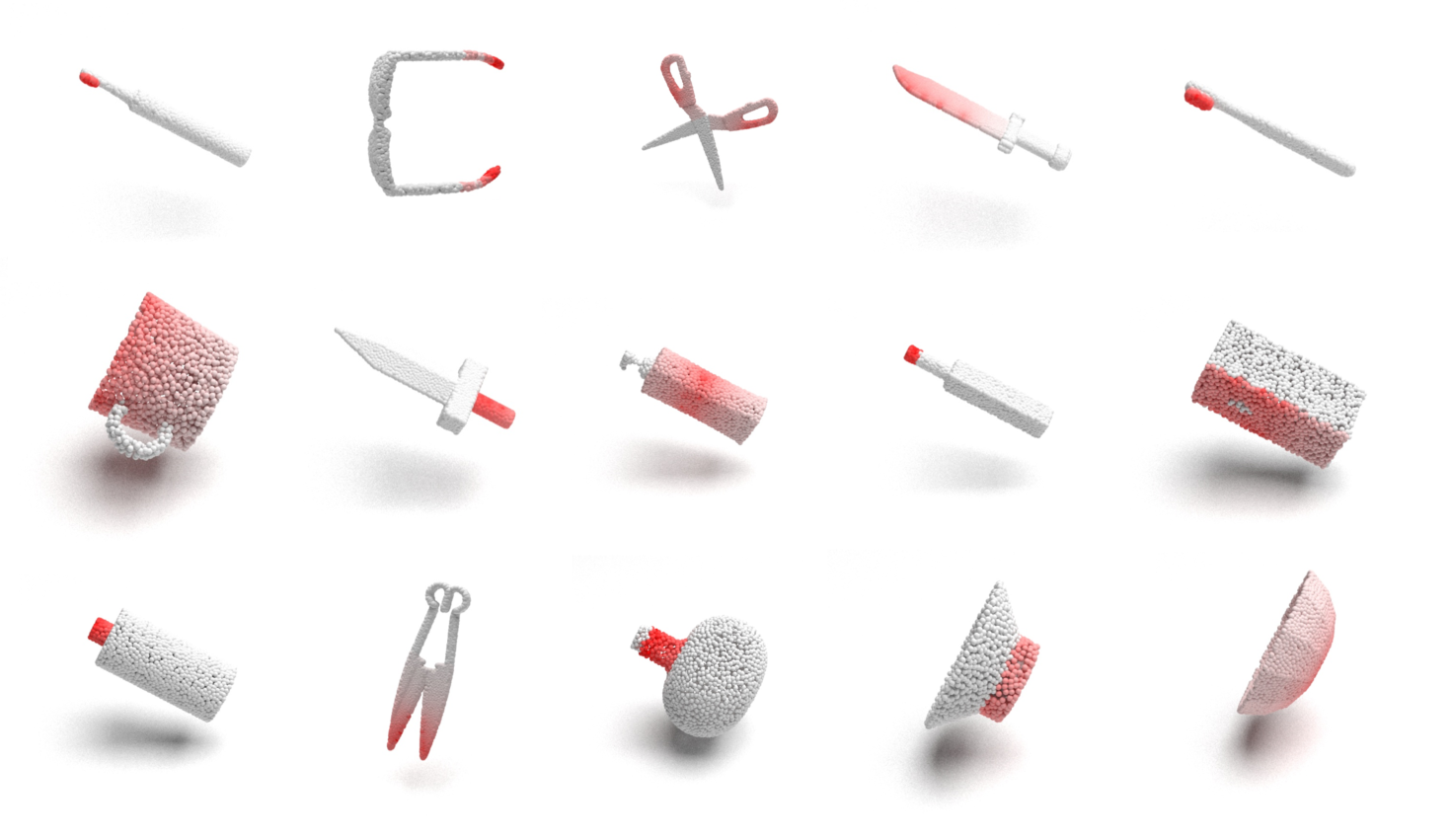}
    \caption{
    \textbf{Representative annotations in UniAfford-Data.}
    Top: RGB images with pixel-level affordance masks.
    Bottom: Point clouds with point-wise affordance annotations.
    Examples are organized under a shared object--affordance taxonomy
    and do not imply instance-level spatial correspondence.
    }
    \label{fig:app-dataset-vis}
\end{figure}

Figure~\ref{fig:app-dataset-vis} presents representative RGB images with pixel-level affordance masks and point clouds with point-wise affordance annotations. The examples illustrate the two spatial supervision formats organized within UniAfford-Data. Their inclusion under a common taxonomy reflects shared functional semantics rather than instance-level geometric alignment.

\section{Additional Method Details}
\label{app:method}

\subsection{Routing Implementation Details}
\label{app:method:routing}

\paragraph{Response states and routing targets.}
The router operates on contextual response states rather than input modality tokens. At response prediction step $t$, let $h_t$ denote the hidden state used to predict the next target token $x_t$, following the indexing in Section~\ref{sec:method:encoding}. During training, assistant responses contain the generic anchors \texttt{<img-aff>} and \texttt{<pc-aff>} when the corresponding branches have supervision. These anchors are shared across object and affordance categories, specifying branch roles rather than category-specific output codes.

Let $V_t$ indicate whether position $t$ belongs to the valid supervised response span, excluding padding and positions outside that span. For valid positions, the route target is derived from the shifted response target:
\begin{equation}
    y_t^{\mathrm{route}}
    =
    \begin{cases}
        \mathrm{img}, &
        x_t = \texttt{<img-aff>}, \\
        \mathrm{pc}, &
        x_t = \texttt{<pc-aff>}, \\
        \mathrm{text}, &
        \text{otherwise}.
    \end{cases}
    \label{eq:app:route-targets}
\end{equation}
These target labels supervise the routing classifier and are distinct from the predicted assignments $r_t$ used during the forward pass. Anchor targets are excluded from the language modeling loss, while ordinary valid response targets retain language supervision.

\paragraph{Branch availability and routing probabilities.}
Branch availability is determined separately from response-position validity. During training, the image route requires both an RGB image and its affordance mask, while the point-cloud route requires both a point cloud and its point-wise annotation. At inference, availability depends on the corresponding observations rather than annotations; the text route remains available.

Let $z_t=g_r(h_t)$ denote the original router logits and $\widetilde{z}_t$ their availability-masked version, with unavailable branch entries set to $-\infty$. The probabilities $p_t=\mathrm{softmax}(\widetilde{z}_t)$ determine hard assignments $r_t=\arg\max_c p_{t,c}$ and enter the structure losses defined below. The availability mask therefore determines which branches are eligible, whereas $V_t$ identifies the response positions used for supervision and query selection.


\paragraph{Ordered queries, padding, and cardinality.}For each affordance branch $b\in\{\mathrm{img},\mathrm{pc}\}$, valid states with predicted assignment $r_t=b$ are projected and concatenated in autoregressive order to form $Q^b$. Its length $K_b$ is the number of valid positions assigned to that branch.

Since $K_b$ varies across samples, query sequences are padded for batched decoding and accompanied by masks identifying valid entries. These query-padding masks are used by the downstream decoders without changing router logits or predicted assignments. They are distinct from the response-validity mask, which selects eligible sequence positions, and the branch-availability mask, which selects eligible prediction branches. 

During training, multi-query outputs are strictly matched one-to-one with their corresponding training annotations based on positional order. Conversely, in zero-query scenarios—where a valid affordance annotation exists but the router fails to correctly identify the placeholder position as an affordance query—the route loss heavily penalizes this misclassification. To ensure training stability, the hidden state at the placeholder position is forced to revert to the downstream decoder as a fallback. During inference, however, these completion mechanisms are disabled: the downstream decoders are dynamically activated for the exact number of queries predicted by the router, without enforced alignment or fallback completion.

\paragraph{Inference interface.}
During autoregressive inference, the learned router predicts branch roles directly from contextual response states. States assigned to an affordance branch provide semantic queries for its dense decoder; task dispatch does not require recognizing a predefined marker generated by the language head. This interface separates branch assignment from prescribed token generation while retaining a shared MLLM context for language and dense prediction.

\subsection{Training Objectives}
\label{app:method:loss}

UniAfford combines language modeling, 2D affordance prediction, 3D affordance prediction, and token routing supervision. Routing objectives train branch assignment, while dense prediction objectives supervise selected representations through the corresponding decoders. We use $L$ for the number of response prediction positions, consistent with the main text.

\paragraph{Language modeling loss.}
Language modeling supervises ordinary valid response targets while excluding routing anchors. Let
$\mathcal{A}=\{\texttt{<img-aff>},\texttt{<pc-aff>}\}$
denote the anchor set and define
$M_t^{\mathrm{txt}}=V_t\,\mathbf{1}[x_t\notin\mathcal{A}]$.
The masked autoregressive objective is
\begin{equation}
    \mathcal{L}_{\mathrm{txt}}
    =
    \frac{
        \sum_{t=1}^{L}
        M_t^{\mathrm{txt}}
        \mathrm{CE}(\widehat{x}_t,x_t)
    }{
        \sum_{t=1}^{L} M_t^{\mathrm{txt}}
    },
    \label{eq:app:language-loss}
\end{equation}
where $\widehat{x}_t$ denotes vocabulary logits produced by the language head from $h_t$. The mask depends on the target token and response validity, not the predicted route. Ordinary response targets therefore retain language supervision even when routing predictions are incorrect, whereas anchor targets provide route supervision without contributing to the language modeling objective.

\paragraph{2D affordance loss.}
For image-space prediction, we combine focal and Dice losses:
\begin{equation}
    \mathcal{L}_{\mathrm{2D}}
    =
    \lambda_f
    \mathcal{L}_{\mathrm{focal}}
    (\widehat{Y}^{2D},Y^{2D})
    +
    \lambda_d
    \mathcal{L}_{\mathrm{dice}}
    (\widehat{Y}^{2D},Y^{2D}),
    \label{eq:app:image-loss}
\end{equation}
where $\widehat{Y}^{2D}$ denotes pixel-wise output logits and $Y^{2D}$ is the corresponding affordance annotation. The focal term emphasizes difficult foreground and background predictions, while the Dice term encourages region-level overlap. This objective is activated only when the image and its target annotation are available, supervising routed image representations through the 2D decoder.

\paragraph{3D affordance loss.}
For point-cloud prediction, we combine binary cross-entropy and Dice losses:
\begin{equation}
    \mathcal{L}_{\mathrm{3D}}
    =
    \lambda_b
    \mathcal{L}_{\mathrm{bce}}
    (\widehat{Y}^{3D},Y^{3D})
    +
    \lambda_{pd}
    \mathcal{L}_{\mathrm{dice}}
    (\widehat{Y}^{3D},Y^{3D}),
    \label{eq:app:point-loss}
\end{equation}
where $\widehat{Y}^{3D}$ contains point-wise affordance logits and $Y^{3D}$ provides the corresponding targets. Binary cross-entropy supervises individual point predictions, while the Dice term encourages overlap with the annotated affordance region. The objective is activated when the point cloud and its annotation are available. For semantically paired samples, each branch uses the spatial annotation associated with its own observation.

\paragraph{Token-level routing loss.}
Let $M_t^{\mathrm{route}}=V_t$ indicate valid positions for route supervision. The routing classifier is trained using the targets from Equation~\ref{eq:app:route-targets}:
\begin{equation}
    \mathcal{L}_{\mathrm{route}}
    =
    \frac{
        \sum_{t=1}^{L}
        M_t^{\mathrm{route}}
        \mathrm{CE}(z_t,y_t^{\mathrm{route}})
    }{
        \sum_{t=1}^{L} M_t^{\mathrm{route}}
    },
    \label{eq:app:route-loss}
\end{equation}
where $z_t$ denotes the original router logits. This position-masked cross-entropy supervises route classification, while branch-availability masking is applied separately when forming $\widetilde{z}_t$ and $p_t$ for hard assignments and structure losses.

\paragraph{Existence and sparsity objectives.}
To regularize branch coverage and query allocation under heterogeneous supervision, we introduce two structure losses on the soft routing probabilities. Let
$\Omega_r=\{t\mid M_t^{\mathrm{route}}=1\}$
contain all valid route-supervision positions, including positions with text-route targets. For each affordance branch $b\in\{\mathrm{img},\mathrm{pc}\}$, we compute
\begin{equation}
    a^b
    =
    1-\prod_{t\in\Omega_r}(1-p_{t,b}),
    \qquad
    c^b
    =
    \sum_{t\in\Omega_r}p_{t,b},
    \label{eq:app:route-statistics}
\end{equation}
where $a^b$ is a noisy-or estimate of branch activation and $c^b$ is the expected token count under the soft routing distributions. These quantities provide differentiable estimates of whether a branch receives queries and how much routing probability is allocated to it.

The existence and sparsity objectives are
\begin{equation}
    \begin{aligned}
        \mathcal{L}_{\mathrm{exist}}
        &=
        \sum_{b\in\{\mathrm{img},\mathrm{pc}\}}
        \mathrm{BCE}(a^b,y^b_{\mathrm{avail}}), \\
        \mathcal{L}_{\mathrm{sparse}}
        &=
        \sum_{b\in\{\mathrm{img},\mathrm{pc}\}}
        \mathrm{SmoothL1}(c^b,\tau^b),
    \end{aligned}
    \label{eq:app:structure-losses}
\end{equation}
where $y^b_{\mathrm{avail}}$ indicates whether branch $b$ has both an observation and its annotation. The target $\tau^b$ is a small positive token count for available branches and zero otherwise. The existence term encourages coverage of supervised branches, while the sparsity term discourages redundant query allocation. Both operate on soft probabilities rather than imposing a fixed number of hard-selected queries. Structure losses are computed per sample and averaged over the mini-batch.

The complete routing objective is
\begin{equation}
    \mathcal{L}_{\mathrm{router}}
    =
    \lambda_r\mathcal{L}_{\mathrm{route}}
    +
    \lambda_e\mathcal{L}_{\mathrm{exist}}
    +
    \lambda_s\mathcal{L}_{\mathrm{sparse}},
    \label{eq:app:router-objective}
\end{equation}
To balance the optimization process, the loss components are scaled empirically based on their initial gradient magnitudes during preliminary experiments. Specifically, the main routing objective is assigned a weight of 1.0 ($\lambda_{\mathrm{r}} = 1.0$). For auxiliary routing regularizations, the route existence loss is scaled by 0.5 ($\lambda_{\mathrm{e}} = 0.5$), and the route sparsity loss is assigned a small weight of 0.01 ($\lambda_{\mathrm{s}} = 0.01$). This minimal sparsity weight acts as a gentle structural regularizer to prevent trivial routing collapse without dominating the primary multimodal alignment gradients. 

\paragraph{Overall objective.}
The full objective combines language, spatial, and routing supervision:
\begin{equation}
    \mathcal{L}
    =
    \lambda_{\mathrm{txt}}\mathcal{L}_{\mathrm{txt}}
    +
    m_{\mathrm{2D}}\mathcal{L}_{\mathrm{2D}}
    +
    m_{\mathrm{3D}}\mathcal{L}_{\mathrm{3D}}
    +
    \mathcal{L}_{\mathrm{router}},
    \label{eq:app:overall-objective}
\end{equation}
where $m_{\mathrm{2D}}=y^{\mathrm{img}}_{\mathrm{avail}}$ and
$m_{\mathrm{3D}}=y^{\mathrm{pc}}_{\mathrm{avail}}$ indicate the availability of each observation and its annotation. Image-only and point-cloud-only samples activate their corresponding dense objectives, while semantically paired samples with both annotations activate both. These objectives connect heterogeneous pixel-level and point-level supervision through shared MLLM representations.

\section{Additional Experimental Details}
\label{app:exp:setup}

\subsection{Evaluation Protocols}
\label{app:exp:protocols}

We use two complementary protocols to examine cross-dataset generalization and standalone branch performance. The mixed-training protocol evaluates a jointly trained model, whereas the modality-isolated protocol separately trains and evaluates each branch of the unified architecture.

\paragraph{Mixed-training OOD zero-shot protocol.}
UniAfford is trained on UniAfford-Data and directly evaluated on target benchmarks without target-specific training or fine-tuning. For 2D transfer, we evaluate on AGD20K. For 3D transfer, GEAL* denotes the evaluation subset of LASO-C from the GEAL corruption benchmark, comprising Scale, Jitter, and Rotate perturbations at severity level 2. This protocol tests whether heterogeneous pixel-level and point-level supervision supports transferable affordance understanding across benchmark distributions.

Here, zero-shot refers to cross-dataset transfer without target-specific adaptation, rather than requiring every target object or affordance category to be absent from the training taxonomy. LASO~\citep{LASO} and GEAL~\citep{GEAL} reference results in Table~\ref{tab:mixed-transfer} are presented separately and excluded from the zero-shot rankings.

\paragraph{Modality-isolated protocol.}
Each branch is trained and evaluated using its target visual modality and task instructions. The 2D branch uses RGB images with pixel-level affordance masks and follows the Affordance-R1~\citep{AffordanceR1} protocol on ReasonAff. The 3D branch uses point clouds with point-wise annotations, training on the corresponding PIAD or PIADv2 training split and evaluating on PIAD Unseen or PIADv2 Unseen-OBJ, respectively.

Object and affordance labels specify the grounding task without providing an additional visual modality. These separately trained models assess the standalone capacity of the unified architecture, complementing the jointly trained model's cross-dataset evaluation.

\paragraph{Shared-subset ablation protocol.}
The component ablations in Section~\ref{sec:exp:ablation} and the extended baseline comparisons in Appendix~\ref{app:extended_ablation} use the same fixed UniAfford-Data subset and held-out partition. Single-branch variants use the corresponding modality's training samples, while the full model uses heterogeneous 2D--3D supervision. Dataset organization is described in Appendix~\ref{app:dataset:splits}.

\subsection{Evaluation Metrics}
\label{app:exp:metrics}

\paragraph{2D region-overlap metrics.}
For 2D prediction, we report \textit{gIoU}, \textit{cIoU}, $P_{50}$, and $P_{50\text{--}95}$. Let $\widehat{B}_i$ and $B_i$ denote the predicted and ground-truth foreground masks used for evaluation after mask preprocessing. These evaluation masks are distinct from the decoder logits defined in the method section. For $N_{\mathrm{img}}$ test images, \textit{gIoU} averages image-level intersection-over-union values:
\begin{equation}
    \textit{gIoU}
    =
    \frac{1}{N_{\mathrm{img}}}
    \sum_{i=1}^{N_{\mathrm{img}}}
    \frac{|\widehat{B}_i \cap B_i|}
         {|\widehat{B}_i \cup B_i|}.
    \label{eq:app:giou}
\end{equation}
In contrast, \textit{cIoU} aggregates intersections and unions before computing their ratio:
\begin{equation}
    \textit{cIoU}
    =
    \frac{
        \sum_{i=1}^{N_{\mathrm{img}}}
        |\widehat{B}_i \cap B_i|
    }{
        \sum_{i=1}^{N_{\mathrm{img}}}
        |\widehat{B}_i \cup B_i|
    }.
    \label{eq:app:ciou}
\end{equation}
Thus, \textit{gIoU} assigns equal weight to individual image-level IoUs, whereas \textit{cIoU} measures overlap across accumulated evaluation regions.

$P_{50}$ is the percentage of predictions whose IoU exceeds 0.50. More generally, for an overlap threshold $\tau$,
\begin{equation}
    P_{\tau}
    =
    \frac{100}{N_{\mathrm{img}}}
    \sum_{i=1}^{N_{\mathrm{img}}}
    \mathbf{1}\!\left[
        \operatorname{IoU}(\widehat{B}_i,B_i)>\tau
    \right].
    \label{eq:app:precision-iou}
\end{equation}
$P_{50\text{--}95}$ averages this percentage over evaluation thresholds from 0.50 to 0.95. These thresholds assess overlap after foreground masks have been constructed and are distinct from the score thresholds used to binarize prediction maps.

\paragraph{2D distribution metrics.}
For saliency-style transfer evaluation, we additionally report Kullback--Leibler divergence (\textit{KLD}) and similarity (\textit{SIM}). KLD measures the discrepancy between predicted and ground-truth affordance distributions, while SIM measures the histogram intersection between normalized prediction and target maps. Lower KLD and higher SIM indicate better distributional agreement.

\paragraph{3D affordance grounding metrics.}
For point-cloud prediction, we report Area Under the Curve (\textit{AUC})~\citep{AUC}, mean IoU (\textit{mIoU})~\citep{mIOU}, similarity (\textit{SIM})~\citep{SIM}, and Mean Absolute Error (\textit{MAE})~\citep{MAE}. AUC evaluates the ranking of point-wise affordance scores. The reported mIoU averages region-overlap scores over multiple binarization thresholds, while SIM compares normalized point-wise affordance distributions. MAE measures the average absolute difference between predicted scores and target labels. Lower MAE is better; higher values are better for the other three metrics.

These metrics characterize complementary aspects of grounding: score ranking, thresholded region overlap, distribution agreement, and point-wise error. Reporting all four provides a broader assessment than relying on a single measure, particularly when prediction extent and score distribution affect the metrics differently.

\subsection{Implementation Details}
\label{app:exp:implementation}

\paragraph{Backbone adaptation and optimization.}
We instantiate the shared MLLM with Qwen3-VL and fine-tune its attention and MLP projection layers using LoRA~\citep{LoRA}. The LoRA rank is 8, the scaling factor is 16, and the dropout rate is 0.05. We optimize the model with AdamW~\citep{AdamW}, using linear warmup followed by cosine learning-rate decay. Experiments are conducted on NVIDIA B200 GPUs.

\paragraph{Trainable and frozen modules.}
To ensure efficient training, only a lightweight subset of parameters is optimized. Since parameter-efficient fine-tuning is adopted by default, the original backbone weights of the shared MLLM—including the language modeling head tied to \texttt{embed\_tokens}—remain frozen, with optimization restricted to its injected LoRA adapters. Furthermore, the visual and geometric encoders, specifically the Qwen3-VL SigLIP-based vision tower, the SAM image encoder, and the pretrained SONATA point-cloud encoder, are kept strictly frozen throughout training. Consequently, active parameter updates are confined to the LoRA adapters, the routing projection layers, the 2D and 3D dense decoders, and the whole Router.

\paragraph{Module-specific learning rates.}
We assign separate learning rates to the MLLM adapters, routing head, and dense decoders to control their updates during joint optimization. Table~\ref{tab:app-training-config} summarizes the settings for each evaluation protocol. Unless otherwise specified, images are resized to $1024\times1024$ and point clouds are sampled to 2048 points.

\begin{table}[ht]
    \centering
    \caption{
    Protocol-specific training configurations.
    The MLLM learning rate applies to its LoRA parameters.
    }
    \label{tab:app-training-config}
    \setlength{\tabcolsep}{5pt}
    \resizebox{\linewidth}{!}{
    \begin{tabular}{lcccccc}
        \toprule
        Protocol & Image Resolution & Points & MLLM LR
        & 2D Decoder LR & 3D Decoder LR & Router LR \\
        \midrule
        Mixed-training OOD zero-shot
        & $1024\times1024$ & 2048 & 1e-5
        & 5e-6 & 5e-4 & 1e-3 \\

        2D modality-isolated
        & $1024\times1024$ & -- & 1e-5
        & 1e-5 & -- & 1e-3 \\

        3D modality-isolated
        & -- & 2048 & 1e-5
        & -- & 1e-4 & 1e-3 \\
        \bottomrule
    \end{tabular}
    }
\end{table}

\paragraph{Baseline implementation.}
We follow official baseline protocols where available, using released checkpoints or retraining models with their prescribed data preparation and training recipes. Methods are evaluated using the same metric definitions on the corresponding test splits. For 3D baselines whose architectures require auxiliary non-point-cloud cues, we retain these inputs under their official settings, while UniAfford uses point clouds as its only visual input in the modality-isolated protocol.

The cross-dataset comparison evaluates complete systems under their reported training configurations. Internal ablations and shared-subset comparisons separately examine routing, joint supervision, and decoder coupling under the corresponding controlled settings.

\subsection{Language-Head Diagnostics of Routed States}
\label{app:diagnostic}

\paragraph{Diagnostic procedure.}
We analyze the linguistic content of contextual states used for dense prediction by projecting them through the MLLM language head and inspecting their highest-scoring vocabulary tokens. The analysis focuses on states selected for the image or point-cloud branches. These readouts characterize the information contained in routed representations; branch assignments themselves are predicted by the router rather than determined by decoded token identities.

During training, routing-anchor targets are excluded from the language modeling loss, while selected representations receive pixel-level or point-level supervision through the corresponding decoders. The diagnostic examines the object- and interaction-related information readable from states shaped by the shared MLLM and dense prediction objectives.

\begin{figure}[ht]
    \centering
    \begin{subfigure}{0.48\linewidth}
        \centering
        \includegraphics[width=\linewidth]{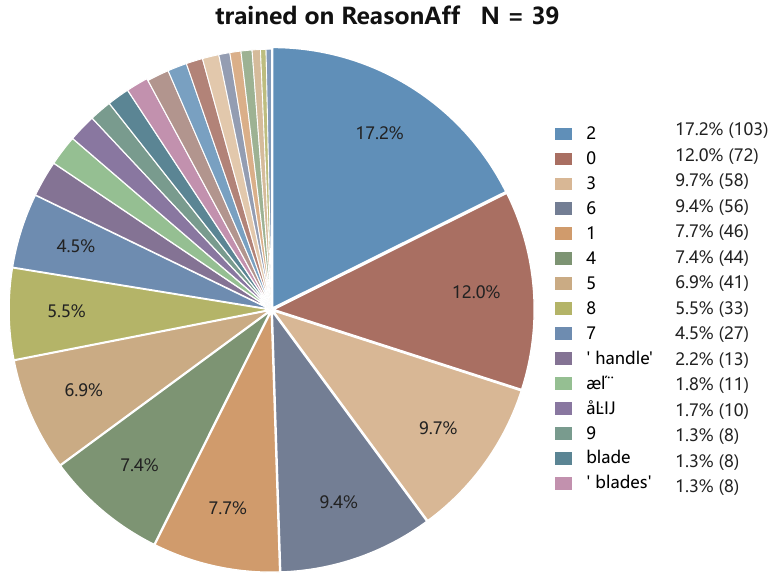}
        \caption{ReasonAff training and evaluation}
        \label{fig:token_pie_ReasonAff}
    \end{subfigure}\hfill
    \begin{subfigure}{0.48\linewidth}
        \centering
        \includegraphics[width=\linewidth]{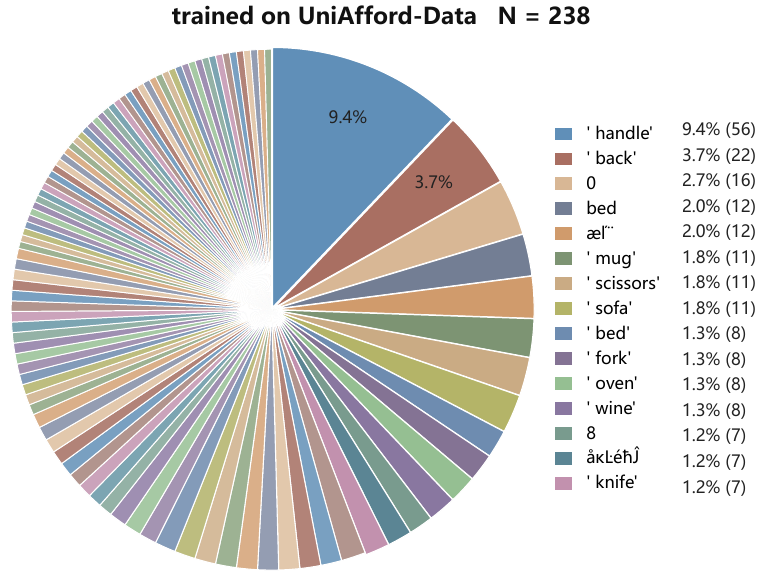}
        \caption{UniAfford-Data model on ReasonAff}
        \label{fig:token_pie_0shot-ReasonAff}
    \end{subfigure}

    \par\medskip

    \begin{subfigure}{0.48\linewidth}
        \centering
        \includegraphics[width=\linewidth]{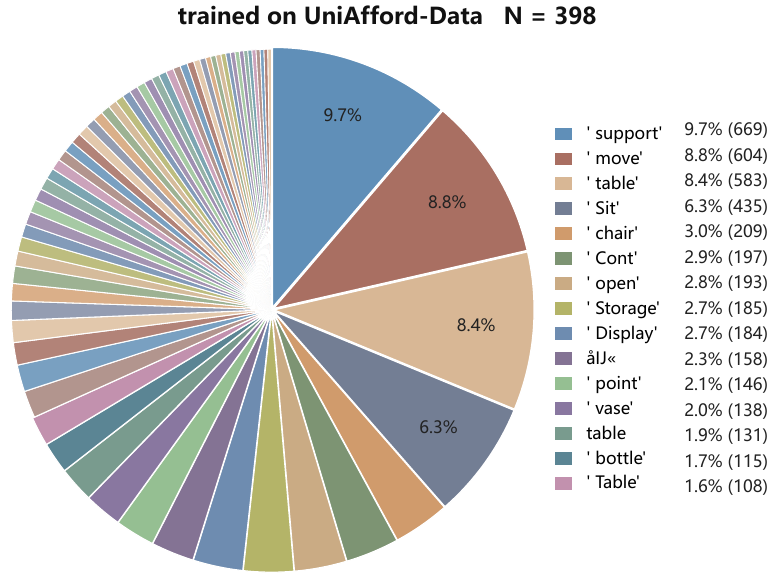}
        \caption{UniAfford-Data model on GEAL}
        \label{fig:token_pie_GEAL}
    \end{subfigure}\hfill
    \begin{subfigure}{0.48\linewidth}
        \centering
        \includegraphics[width=\linewidth]{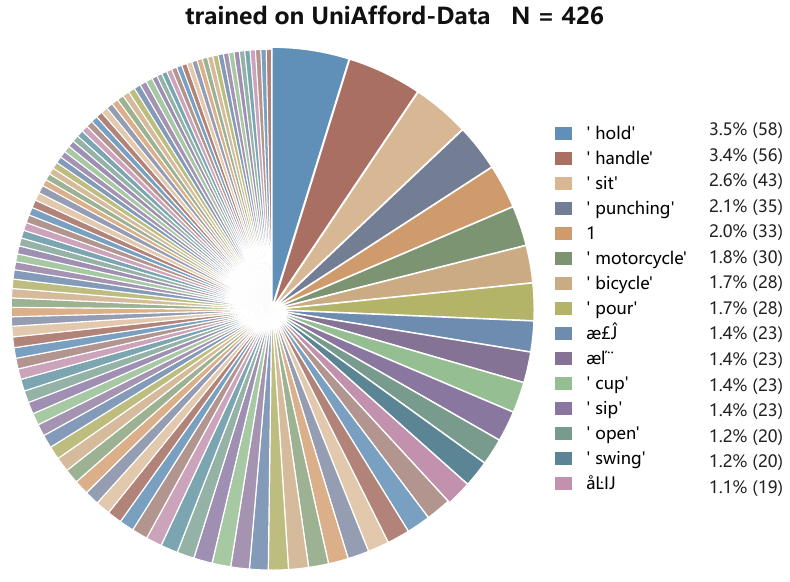}
        \caption{Zero-shot transfer on AGD20K}
        \label{fig:token_pie_0shot-AGD20k}
    \end{subfigure}

    \caption{
    \textbf{Language-head readouts of routed representations.}
    Each panel summarizes decoded-token frequencies for the indicated
    training and evaluation setting. $N$ denotes the number of distinct
    decoded-token categories recorded in the corresponding diagnostic run.
    }
    \label{fig:token_pie_comparison}
\end{figure}

\paragraph{Decoded distributions and semantic content.}
Figure~\ref{fig:token_pie_comparison} compares training on ReasonAff alone with joint training on UniAfford-Data. The ReasonAff-only setting produces a concentrated decoded distribution with $N=39$ categories and frequent numeric tokens. The jointly trained model produces broader reported distributions, with $N=238$ categories on ReasonAff, $N=398$ on GEAL, and $N=426$ on AGD20K.

The UniAfford-Data model's readouts include object- and part-related terms such as `` table'', `` handle'', and `` motorcycle'', together with interaction terms such as `` support'', `` move'', `` hold'', and `` sit''. These observations reveal meaningful object--affordance semantic content in states used for spatial prediction. The ReasonAff panel compares representation diagnostics under different training settings and is not presented as an additional cross-dataset zero-shot result.

\paragraph{Interpretation.}
The language-head readouts demonstrate that routed states retain linguistically accessible functional information while serving downstream dense prediction. Together with the learned-routing versus fixed-anchor comparison, they support the use of contextual MLLM states as semantic interfaces for affordance decoding.

The category counts describe the observed readout distributions rather than a sample-size-controlled measure of semantic richness. Numeric readouts alone do not imply that a state lacks useful information, and vocabulary inspection characterizes representation content rather than the causal basis of routing decisions.

We acknowledge that a minority of the routed tokens still decode into semantically irrelevant characters. We attribute this phenomenon to two primary factors: first, legacy artifacts originating from the MLLM’s prior training distribution; and second, representation drift induced by the downstream decoders, which pull the token embeddings away from the pure language space to better align with the spatial requirements of dense prediction during joint training. Ultimately, this striking contrast definitively highlights that the scale and diversity of the training dataset are directly correlated with the richness of the learned semantics. Our unified heterogeneous training on large-scale data prevents the router from overfitting to template neighborhoods, compelling the network to utilize genuine, high-level object–-affordance semantics as a universal interface for cross-modal dense prediction.

\subsection{Extended Ablations and Cross-Modal Learning Gains}
\label{app:extended_ablation}

\paragraph{Shared-subset baseline comparison.}
We extend the component analysis by training representative baselines on the same UniAfford-Data subset used in Section~\ref{sec:exp:ablation}. The 2D baseline is the AffordanceNet framework associated with RAGNet~\citep{RAGNet}, and the 3D baseline is GREAT~\citep{GREAT}. Each baseline uses the corresponding modality from the shared data subset, while the full UniAfford model jointly uses 2D and 3D supervision. Evaluation follows the same held-out partition for the relevant prediction space.

\begin{table}[ht]
    \centering
    \caption{
    Performance on the shared UniAfford-Data ablation subset.
    Specialized baselines use their respective modalities, whereas
    the full UniAfford model uses heterogeneous 2D--3D supervision.
    }
    \label{tab:app-baseline-retrain}
    \setlength{\tabcolsep}{5pt}
    \resizebox{\linewidth}{!}{
    \begin{tabular}{lcccccc}
        \toprule
        \multirow{2}{*}{Method}
        & \multicolumn{2}{c}{2D Metrics}
        & \multicolumn{4}{c}{3D Metrics} \\
        \cmidrule(lr){2-3} \cmidrule(lr){4-7}
        & gIoU$\uparrow$ & cIoU$\uparrow$
        & AUC$\uparrow$ & mIoU$\uparrow$
        & SIM$\uparrow$ & MAE$\downarrow$ \\
        \midrule
        AffordanceNet~\citep{RAGNet}
        & 31.44 & 21.37 & -- & -- & -- & -- \\

        GREAT~\citep{GREAT}
        & -- & -- & 71.02 & 13.14 & 0.415 & 0.154 \\
        \midrule
        \textbf{UniAfford}
        & \textbf{68.79} & \textbf{58.36}
        & \textbf{84.43} & \textbf{34.56}
        & \textbf{0.583} & \textbf{0.105} \\
        \bottomrule
    \end{tabular}
    }
\end{table}

\paragraph{Performance under shared 2D training data.}
The 2D-only UniAfford variant in Table~\ref{tab:ablation} achieves 41.41 gIoU, compared with 31.44 for the RAGNet baseline trained on the same 2D data subset. This 9.97-point improvement is achieved without additional 3D supervision for that variant, establishing a complete-system advantage under shared 2D training data. The specific contribution of learned routing is evaluated separately through the matched-architecture comparison below.

\paragraph{Joint supervision with unchanged branch backbones.}
Joint training further raises 2D gIoU from 41.41 to 68.79 and 3D mIoU from 30.07 to 34.56 relative to the corresponding single-branch variants. The 3D-only and joint variants use the same SONATA backbone configuration, so the 4.49-point mIoU improvement does not require replacing the geometric backbone with a larger model.

These gains establish the additional value of heterogeneous supervision within UniAfford. Pixel-level and point-level annotations jointly improve grounding through the shared architecture, directly supporting the motivation to study 2D and 3D affordances within a common learning framework.

\paragraph{Learned routing under matched architecture and data.}
The fixed-anchor comparison in Table~\ref{tab:ablation} retains the same backbones and dense decoders while replacing learned routing with anchor-based state selection. Learned routing improves 2D gIoU from 62.28 to 68.79 and 3D mIoU from 17.46 to 34.56, demonstrating its advantage over fixed-anchor selection within the proposed architecture.

Together, these experiments provide complementary evidence for system performance under shared training data, the benefits of joint supervision with unchanged branch backbones, and the effectiveness of learned state selection under matched architectural components.

\subsection{Detailed Computational Cost Analysis}
\label{app:cost}

\paragraph{Profiling settings.}
We profile UniAfford under the H1 task settings, measuring 2D inference on AGD20K and 3D inference on GEAL*. We report complete-pipeline FLOPs, latency, and throughput, together with separately profiled MLLM, router, and dense-decoder costs. UniAfford uses BF16 computation with FP32 for the 3D encoder and decoder, while baseline models are evaluated in BF16.

\paragraph{2D affordance inference.}
Table~\ref{tab:cost-2d} demonstrates a substantial efficiency advantage over Affordance-R1~\citep{AffordanceR1}. UniAfford requires 19,658.40 GFLOPs per sample, approximately 27.6\% fewer than Affordance-R1, and achieves 8.20 samples/s compared with 1.05 samples/s, corresponding to approximately $7.8\times$ the throughput. Its end-to-end latency is 122.01 ms per sample, while the separately profiled MLLM, router, and 2D decoder require 86.14 ms, 0.08 ms, and 34.37 ms, respectively.

AffordanceNet from RAGNet remains faster at 74.30 ms per sample. The comparison therefore shows UniAfford's efficiency gain over the reasoning-based MLLM baseline alongside its computational cost relative to a specialized 2D model.

\begin{table}[ht]
    \centering
    \caption{
    Computational cost on AGD20K.
    Top-level rows report complete-pipeline measurements; indented rows
    report separately profiled UniAfford components.
    }
    \label{tab:cost-2d}
    \setlength{\tabcolsep}{5pt}
    \resizebox{\linewidth}{!}{
    \begin{tabular}{lccc}
        \toprule
        Method / Component
        & FLOPs (GFLOPs/sample)
        & Latency (ms/sample)
        & Throughput (samples/s) \\
        \midrule
        Affordance-R1~\citep{AffordanceR1}
        & 27,165.82 & 955.44 & 1.05 \\

        AffordanceNet~\citep{RAGNet}
        & 10,315.48 & 74.30 & 13.46 \\

        \textbf{UniAfford}
        & 19,658.40 & 122.01 & 8.20 \\

        \quad $\hookrightarrow$ MLLM
        & 13,691.75 & 86.14 & -- \\

        \quad $\hookrightarrow$ Router
        & 1.38 & 0.08 & -- \\

        \quad $\hookrightarrow$ 2D decoder
        & 5,965.28 & 34.37 & -- \\
        \bottomrule
    \end{tabular}
    }
\end{table}

\paragraph{3D affordance inference.}
Table~\ref{tab:cost-3d} reports the corresponding 3D measurements. The full UniAfford pipeline requires 1,930.87 GFLOPs and 66.27 ms per sample, exceeding the computational cost of the specialized GREAT~\citep{GREAT} and IAGNet~\citep{IAGNet} baselines. The dominant reported component is the MLLM, at 1,903.81 GFLOPs and 55.11 ms.

In comparison, the router requires 1.38 GFLOPs and 0.07 ms, while the 3D decoder requires 7.11 GFLOPs and 3.63 ms. These component measurements show modest routing and decoding overhead relative to the shared MLLM backbone. The breakdown identifies backbone efficiency as an important direction for reducing total inference cost, while the complete-pipeline measurements provide the end-to-end comparison with specialized baselines.

\begin{table}[ht]
    \centering
    \caption{
    Computational cost on GEAL*.
    Top-level rows report complete-pipeline measurements; indented rows
    report separately profiled UniAfford components.
    }
    \label{tab:cost-3d}
    \setlength{\tabcolsep}{5pt}
    \resizebox{\linewidth}{!}{
    \begin{tabular}{lccc}
        \toprule
        Method / Component
        & FLOPs (GFLOPs/sample)
        & Latency (ms/sample)
        & Throughput (samples/s) \\
        \midrule
        GREAT~\citep{GREAT}
        & 21.15 & 6.27 & 159.42 \\

        IAGNet~\citep{IAGNet}
        & 13.68 & 8.22 & 121.69 \\

        \textbf{UniAfford}
        & 1,930.87 & 66.27 & 15.09 \\

        \quad $\hookrightarrow$ MLLM
        & 1,903.81 & 55.11 & -- \\

        \quad $\hookrightarrow$ Router
        & 1.38 & 0.07 & -- \\

        \quad $\hookrightarrow$ 3D decoder
        & 7.11 & 3.63 & -- \\
        \bottomrule
    \end{tabular}
    }
\end{table}

\paragraph{GPU memory footprint.}
Table~\ref{tab:cost-vram} summarizes the single-GPU memory footprint of UniAfford evaluated under a standard profiling configuration with a batch size of 1. Allocated memory is approximately 10.26 GiB upon model loading, reaching an inference peak of 11.60 GiB, with a peak reserved memory of 13.78 GiB. We emphasize that these empirical measurements are reported for reference under this specific batch-1 setup. 

\begin{table}[ht]
    \centering
    \caption{Single-GPU memory footprint of UniAfford.}
    \label{tab:cost-vram}
    \begin{tabular}{@{}p{0.35\linewidth}cp{0.43\linewidth}@{}}
        \toprule
        Memory Metric & Size (GiB) & Description \\
        \midrule
        Model baseline (allocated)
        & $\sim$10.26
        & Allocated memory after loading model weights \\

        Inference peak (allocated)
        & $\sim$11.60
        & Peak allocated memory during inference \\

        Inference peak (reserved)
        & $\sim$13.78
        & Peak memory reserved by the PyTorch allocator \\
        \bottomrule
    \end{tabular}
\end{table}

\section{Additional Qualitative Results and Analysis}
\label{app:qualitative}

\begin{figure}[ht]
    \centering
    \includegraphics[width=\linewidth]{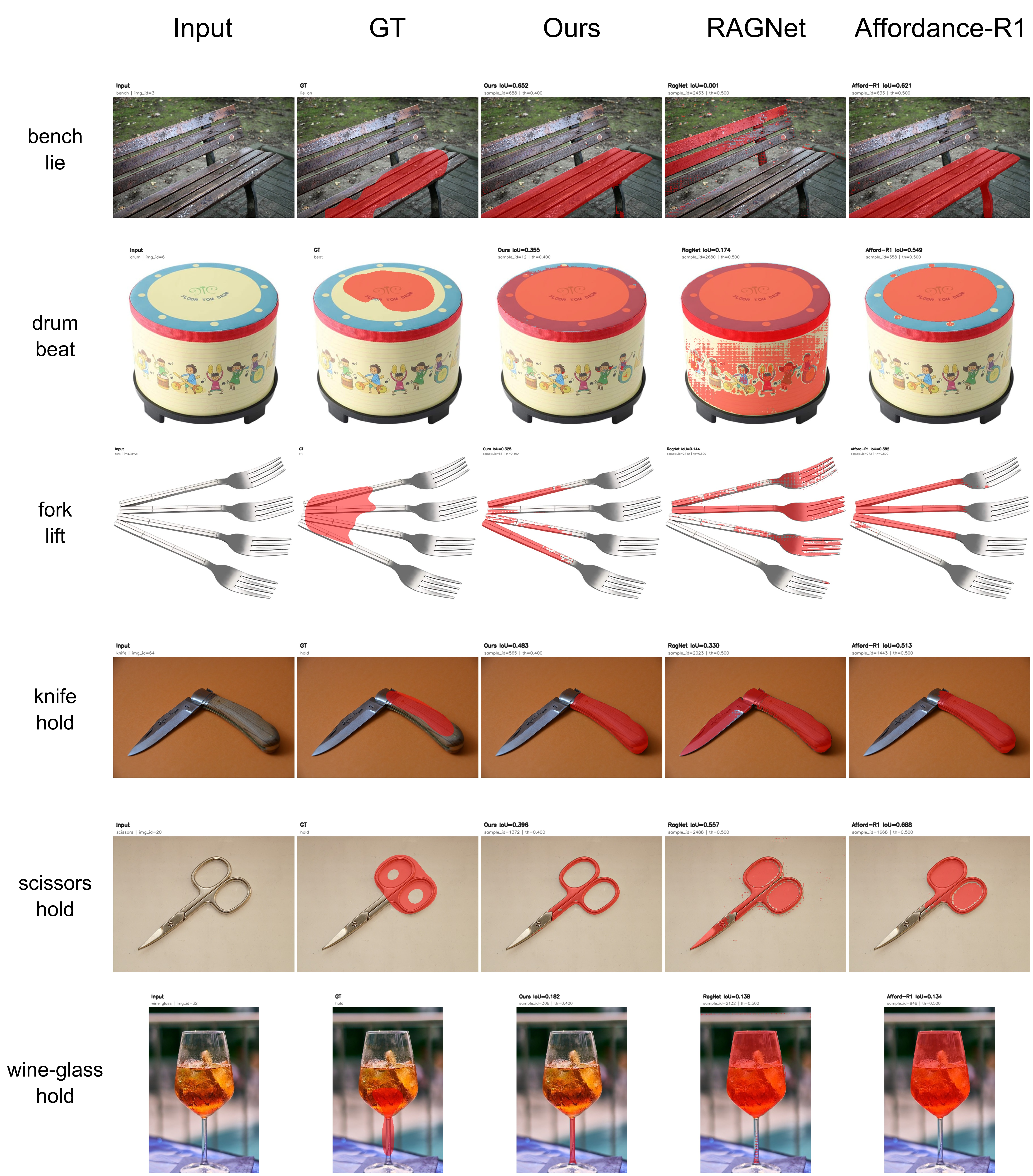}
    \caption{
    \textbf{Qualitative 2D zero-shot comparisons on AGD20K.}
    Columns show the input image, ground-truth annotation, UniAfford,
    the RAGNet baseline, and Affordance-R1.
    Red overlays visualize annotated or predicted affordance regions.
    }
    \label{fig:app_2d_compared_res}
\end{figure}

\subsection{2D Zero-Shot Qualitative Results on AGD20K}
\label{app:2d_qualitative}

Figure~\ref{fig:app_2d_compared_res} presents zero-shot affordance predictions on AGD20K, comparing UniAfford with the 2D baseline from RAGNet~\citep{RAGNet} and Affordance-R1~\citep{AffordanceR1}. Following the H1 protocol, UniAfford is trained on UniAfford-Data without target-specific fine-tuning. These comparisons highlight fine-grained functional localization under cross-dataset distribution shifts and provide a spatial interpretation of the quantitative transfer results.

\paragraph{Instruction-conditioned spatial selectivity.}
UniAfford produces spatially concentrated masks in the illustrated tool-interaction examples. For ``scissors hold'', its prediction focuses on the regions relevant to holding while remaining separated from the cutting blades. Compared with the displayed annotation, the prediction follows a narrower interaction region, illustrating a difference in spatial extent under cross-dataset transfer. For ``knife hold'', UniAfford similarly concentrates on the graspable region instead of extending across the full object. In comparison, the RAGNet predictions in these examples extend further into the blade regions. Overall, these qualitative comparisons illustrate differences in instruction-conditioned spatial selectivity and prediction extent among the evaluated methods; aggregate quantitative results are reported in Table~\ref{tab:mixed-transfer}.

\subsection{3D Zero-Shot Qualitative Results on GEAL*}
\label{app:3d_qualitative}

\begin{figure}[ht]
    \centering
    \includegraphics[width=\linewidth]{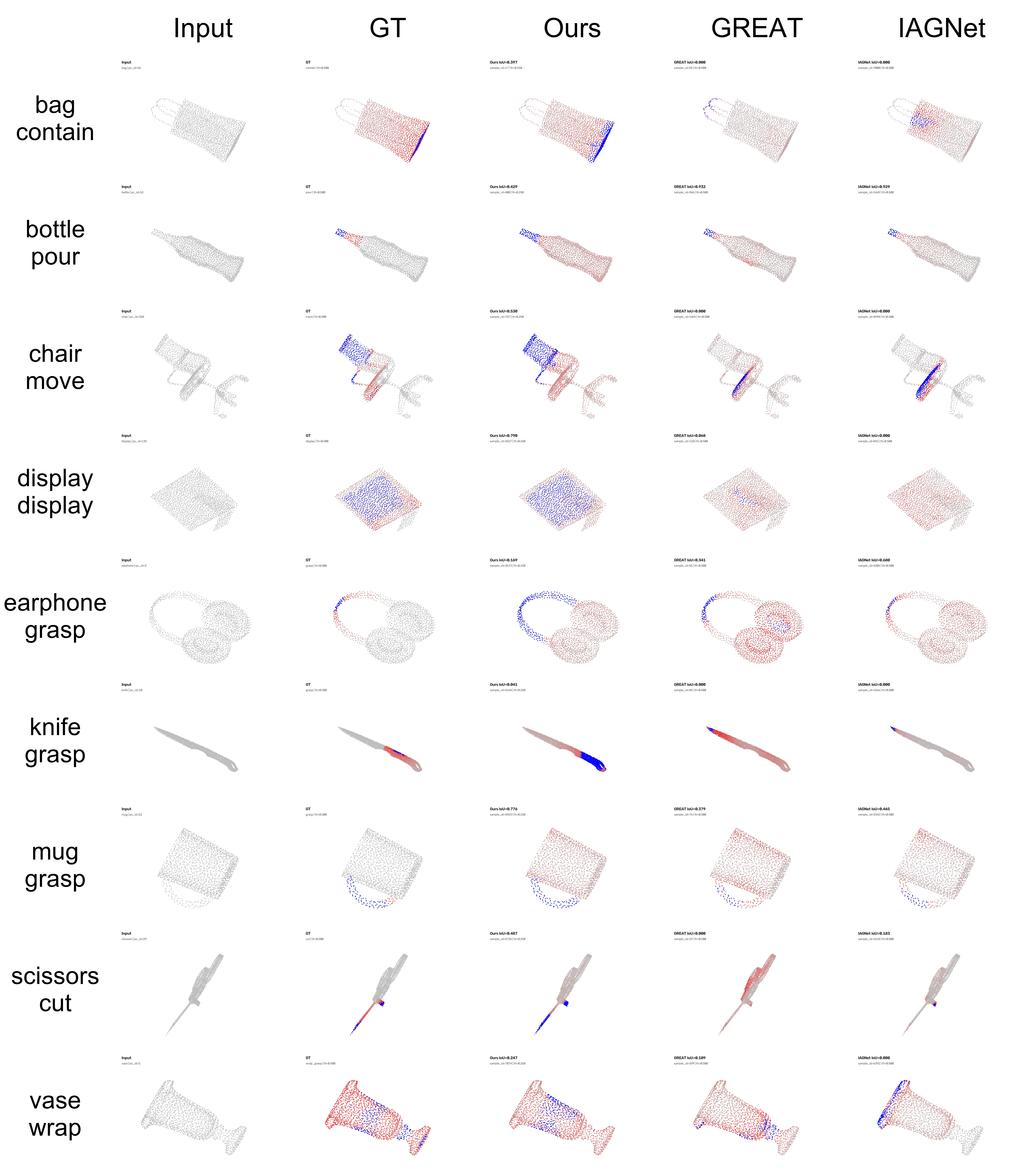}
    \caption{
    \textbf{Qualitative 3D zero-shot comparisons on GEAL*.}
    Columns show the input point cloud, ground-truth annotation, UniAfford, the GREAT baseline, and IAGNet. 
    Red colors represent point-wise affordance scores, with blue highlighting the highest-confidence regions. 
    }
    \label{fig:app_3d_compared_res}
\end{figure}

Figure~\ref{fig:app_3d_compared_res} presents qualitative point-cloud comparisons between UniAfford and representative 3D affordance grounding baselines, GREAT and IAGNet, on GEAL*. The color intensity of the red points represents the predicted affordance score, with darker red indicating a higher response. Blue points highlight the highest-confidence regions under the visualization threshold. 

The comparisons illustrate differences in instruction-conditioned spatial selectivity. In the ``knife grasp'' example, GREAT and IAGNet assign high responses to both the blade and the handle, whereas UniAfford produces a more concentrated response on the region that is more compatible with the requested grasping interaction. Its prediction also exhibits a clearer separation between high- and low-response regions. Similar patterns can be observed in other examples, where UniAfford forms spatially coherent affordance regions while reducing responses on parts that are less relevant to the specified interaction. Across the illustrated examples, UniAfford produces more selective and spatially coherent responses for the specified interactions.

UniAfford does not achieve the highest overlap score in every illustrated example. Nevertheless, in cases such as ``earphone grasp'' and "bottle pour``*'', its predictions remain spatially coherent and cover functionally plausible interaction regions under visual inspection. These examples provide a qualitative view of the predicted affordance distributions and complement the aggregate quantitative results reported in Table~\ref{tab:mixed-transfer}.

\end{document}